\documentclass[10pt,twocolumn,letterpaper]{article}

\usepackage{iccv}
\usepackage{times}
\usepackage{epsfig}
\usepackage{graphicx}
\usepackage{amsmath}
\usepackage{amssymb}
\usepackage{booktabs}

\usepackage[pagebackref,breaklinks,colorlinks]{hyperref}

\usepackage{overpic} 
\usepackage{color}
\usepackage[dvipsnames]{xcolor}
\usepackage{paralist}
\usepackage{soul} 
\definecolor{turquoise}{cmyk}{0.65,0,0.1,0.3}
\definecolor{purple}{rgb}{0.65,0,0.65}
\definecolor{dark_green}{rgb}{0, 0.5, 0}
\definecolor{orange}{rgb}{0.8, 0.6, 0.2}
\definecolor{red}{rgb}{0.8, 0.2, 0.2}
\definecolor{darkred}{rgb}{0.6, 0.1, 0.05}
\definecolor{blueish}{rgb}{0.0, 0.3, .6}
\definecolor{light_gray}{rgb}{0.7, 0.7, .7}
\definecolor{pink}{rgb}{1, 0, 1}
\definecolor{greyblue}{rgb}{0.25, 0.25, 1}

\newif\ifshowcomments
\showcommentstrue 
\ifshowcomments

\newcommand{\myparagraph}[1]{\noindent\textbf{#1.}}
\newcommand{\methodname}{TransFusion\xspace}
\newcommand{\improvetest}{40.4\%\xspace}

\newcommand{\improveval}{25.6\%\xspace}

\usepackage{arydshln}
\usepackage{enumitem} 
\usepackage{multirow}
\usepackage{subcaption}
\usepackage{capt-of}

\usepackage[capitalize]{cleveref}
\crefname{section}{Sec.}{Secs.}
\Crefname{section}{Section}{Sections}
\Crefname{table}{Table}{Tables}
\crefname{table}{Tab.}{Tabs.}

\usepackage{pifont}
\newcommand{\cmark}{\ding{51}}%
\newcommand{\xmark}{\ding{55}}%

\iccvfinalcopy

\makeatletter
\def\ps@myheadings{%
    \let\@oddfoot\@empty\let\@evenfoot\@empty
    \def\@evenhead{\thepage\hfil\slshape\leftmark}%
    \def\@oddhead{{\slshape\rightmark}\hfil\thepage}%
    \let\@mkboth\@gobbletwo
    \let\sectionmark\@gobble
    \let\subsectionmark\@gobble
    }
  \if@titlepage
  \renewcommand\maketitle{\begin{titlepage}%
  \let\footnotesize\small
  \let\footnoterule\relax
  \let \footnote \thanks
  \null\vfil
  \vskip 60\p@
  \begin{center}%
    {\LARGE \@title \par}%
    \vskip 3em%
    {\large
     \lineskip .75em%
      \begin{tabular}[t]{c}%
        \@author
      \end{tabular}\par}%
      \vskip 1.5em%
    {\large \@date \par}
  \end{center}\par
  \@thanks
  \vfil\null
  \end{titlepage}%
  \setcounter{footnote}{0}%
}
\else
\renewcommand\maketitle{\par
  \begingroup
    \renewcommand\thefootnote{\@fnsymbol\c@footnote}%
    \def\@makefnmark{\rlap{\@textsuperscript{\normalfont\@thefnmark}}}%
    \long\def\@makefntext##1{\parindent 1em\noindent
            \hb@xt@1.8em{%
                \hss\@textsuperscript{\normalfont\@thefnmark}}##1}%
    \if@twocolumn
      \ifnum \col@number=\@ne
        \@maketitle
      \else
        \twocolumn[\@maketitle]%
      \fi
    \else
      \newpage
      \global\@topnum\z@   
      \@maketitle
    \fi
    \thispagestyle{plain}\@thanks
  \endgroup
  \setcounter{footnote}{0}%
}
\makeatother

\begin{document}

\title{Summarize the Past to Predict the Future: Natural Language Descriptions of Context Boost Multimodal Object Interaction Anticipation}

\author{ Razvan-George Pasca\textsuperscript{1*} 
\quad
Alexey Gavryushin\textsuperscript{1*}
\quad
Muhammad Hamza\textsuperscript{1}\\
\quad
Yen-Ling Kuo\textsuperscript{2}
\quad
Kaichun Mo\textsuperscript{3}
\quad
Luc Van Gool\textsuperscript{1}
\quad
Otmar Hilliges\textsuperscript{1}
\quad 
Xi Wang\textsuperscript{1}
\\
    \normalsize\textsuperscript{1} ETH Zürich \quad
    \textsuperscript{2} Massachusetts Institute of Technology \quad
    \textsuperscript{3} NVIDIA
}

\twocolumn[{%
\renewcommand\twocolumn[1][]{#1}%
\maketitle
\begin{center}
  \newcommand{\teaserwidth}{\textwidth}
  \vspace{-0.7cm}
  \centerline{\includegraphics[width=0.85\linewidth]{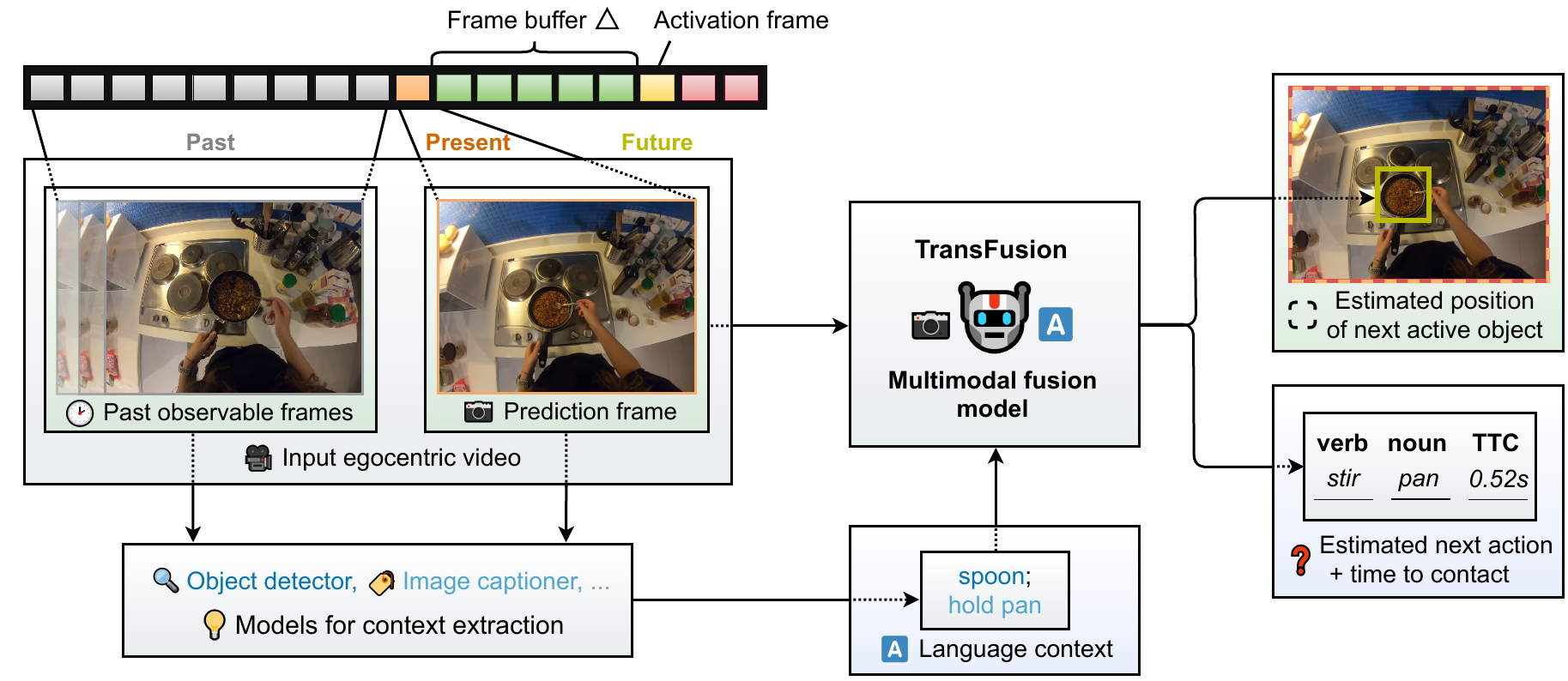}}
    \captionof{figure}{\textbf{TransFusion: Multimodal fusion transformer for short-term object interaction anticipation.} Given a video sequence of past observations, the object interaction anticipation task aims to predict a set of objects visible in the current frame that will be interacted with in the future, i.e. in the activation frame that is $\Delta$ frames away from the current prediction frame. Additionally, the task requires estimating  the bounding box, the associated action described by a verb-noun pair, and the time to contact for each predicted object.
    We propose \methodname, a multimodal fusion architecture that uses a language summaries of past actions to effectively predict future object interactions.      
    }
    \label{fig:Teaser.drawio}
\end{center}
}]
\def\thefootnote{*}\footnotetext{Authors contributed equally.}

\begin{abstract}
We study object interaction anticipation in egocentric videos. This task requires an understanding of the spatio-temporal context formed by past actions on objects, coined \emph{action context}. We propose \methodname, a multimodal transformer-based architecture. It exploits the representational power of language by summarizing the action context. \methodname leverages pre-trained image captioning and vision-language models to extract the action context from past video frames. This action context together with the next video frame is processed by the multimodal fusion module to forecast the next object interaction. Our model enables more efficient end-to-end learning. 
The large pre-trained language models add commonsense and a generalization capability. Experiments on Ego4D and EPIC-KITCHENS-100 show the effectiveness of our multimodal fusion model. They also highlight the benefits of using language-based context summaries in a task where vision seems to suffice. Our method outperforms state-of-the-art approaches 
by \improvetest in relative terms in overall mAP on the Ego4D test set.
We validate the effectiveness of \methodname via experiments on EPIC-KITCHENS-100.
Video and code are available at: \small{\url{https://eth-ait.github.io/transfusion-proj/}.}
\end{abstract}
\vspace{-0.5cm}

\section{Introduction}
\label{sec:intro}
The ability to predict plausible future human-object interactions is important for many assistive technologies. %
The task of \textit{short-term object interaction anticipation}~\cite{next_active_object_intro, ego4d} is defined as predicting \emph{which} object a user will interact with next and \emph{what} action will be performed, given an egocentric video input. 
Providing an effective solution would help artificial agents to assist humans in their daily activities, e.g. care robots.

This task is challenging because it requires reasoning about user intent~\cite{bruner1981intention, orr2002nature, sheeran2002intention}, which is not directly observable.
For example, if the video shows the user holding a vegetable, the most likely next action is to slice it, but only with a knife and cutting board present. Without those, there is significantly more uncertainty. Therefore, understanding what has happened in the past and what relevant objects are present %
is critical for predicting the next object interaction. We collectively refer to these concepts as \emph{action context}.     
Existing interaction forecasting methods do not explicitly model action context and rely on neural networks to extract that information from fixed-sized video chunks~\cite{egonet_action_objects, next_active_object_intro, ego4d, siyu_tang_hand_nao}. 

In this paper, we propose to use language summarization of the past as an explicit representation of action context. This filters out ambiguities caused by visual clutter. It also puts focus on objects and actions directly relevant to the ongoing and likely future activities. For the object interaction anticipation, we next introduce a multimodal fusion approach.
Specifically, we introduce a transformer-based multimodal fusion model, \methodname. 
As shown in \autoref{fig:Teaser.drawio}, it takes language summaries of the action context and the current single frame of a video as input and predicts future object interactions. 
We employ existing image captioning systems to generate a concise description of past actions described by pairs of verbs and nouns. 
We use CLIP~\cite{CLIP} to detect salient objects in the past frames, and deem them task-relevant. 
This vision-language model was trained on pairs of images and captions. 
The verb-noun pairs together with the list of salient objects form the summary of the past action context.
Leveraging pre-trained vision-language models yields more generic representations of the past action context and enables us to better generalize towards various scenarios, in particular those long-tailed ones not well represented in the dataset. 

Our experiments show that our language-based summaries can effectively represent the past action context. 
We use two challenging egocentric datasets, Ego4D~\cite{ego4d} and EPIC-KITCHENS-100 (EK100)~\cite{damen2022rescaling}.
Our model improves interaction prediction accuracy over models pre-trained on large video datasets by \improveval on the Ego4D challenge validation set and \improvetest on the test set. 
\methodname performs similarly
on EK100~\cite{damen2022rescaling}.
Such improvements show that large vision-language models can be useful even for a task that seems purely visual. 

\noindent In summary, our contributions are:
 \begin{compactitem}
    \item The concept of leveraging concise language-based summaries of action context. 
    \item \methodname, a multimodal fusion model for object interaction anticipation that combines action context descriptions and images.
    \item Experiments on Ego4D show improvements over state-of-the-art methods on the object interaction anticipation task, in particular in long-tailed classes.  
 \end{compactitem}  
\section{Related work}

\myparagraph{Object interaction anticipation}
Several works have been proposed for human-object interaction prediction.
Furnari et al. \cite{next_active_object_intro} first introduced \textit{next-active-object} (NAO) prediction, which predicts the active/not active label for each object using their motion trajectories features but it doesn't consider action anticipation.
Closely related to NAO, Bertasius et al. \cite{egonet_action_objects} introduced \textit{action-objects}, which considers objects that capture the human actor's visual attention or tactile interaction.
Bertasius et al. use the data from a stereo camera system in a two-stream network, integrating RGB and depth information. They show that certain aspects of human actions can be predicted by exploiting the spatial configuration of the objects and the actor's head.
Their approach is not easily scalable, because they need stereo. 

In addition to different tasks, Liu et al. \cite{siyu_tang_hand_nao} propose a model that improves NAO prediction by incorporating future hand trajectory modeling.
They obtain impressive results using 3D CNNs, confirming that hand motion cues and scene dynamics have important explanatory power in predicting the short-term future.
Nonetheless, their approach encounters scaling limitations because additional ground-truth labels for hand trajectories are needed.

More recently, the Ego4D dataset \cite{ego4d} proposed \textit{object interaction anticipation}. 
Apart from locating the NAO, the task in Ego4D also requires the prediction of associated nouns and verbs, as well as the time to contact (TTC) for the future interacted objects.
The Ego4D method uses the current frame and past video frames as input and employs a two-stage approach to learn noun and box prediction first and then the verbs and TTC prediction.
Compared to their work, we perform the task in a single pass, by tuning the multimodal fusion module on top of the extracted RGB and language features to learn nouns and verbs jointly, resulting in better performance. 

\myparagraph{Action anticipation}
Besides localizing future object interactions, predicting the next action steps is an equally important task that AI systems need to perform to interact with humans. 
A longer prediction timespan requires a better grasp of the structure behind the succession of actions.
Anticipative Video Transformer (AVT) \cite{anticipative_video_transformer} is one of the best-performing architectures on the EK100 \cite{ek_100} action anticipation benchmark.
AVT uses a pre-trained Visual Transformer \cite{ViT} as the image feature backbone and attends to the previously observed video frames to anticipate future actions. 
MeMViT \cite{memvit} improves AVT by exploiting a longer temporal context, highlighting the importance of modeling a long sequence of previous actions.
Other model architectures such as recurrent neural networks \cite{rolling_unrolling_lstm, imagine_rnn} and multimodal temporal CNNs \cite{temporl_binding_network} are also used to model the temporal contexts.
Note that these methods are designed for action classification using video data and adapting them for object detection is not trivial. 

\myparagraph{Vision and language models}
Our approach combines visual and language modalities to improve object anticipation.
Most recent multimodal architectures are based on Transformer fusion schemes applied on features extracted by various encoders~\cite{frozen_in_time, trans_vg, findit}. We follow the spirit by employing self-attention-based modality fusion. Our work differs in one significant aspect: the language features represent the past action context. This serves as an additional signal to disambiguate the actor's intent given similar visual scenes. The model has to learn the next action and objects from the multimodal fusion of language context summaries and visual cues in a considerably smaller data regime.

More broadly, there are numerous transformer-based works aimed at learning cross-modal representations between vision and language \cite{CLIP, CLIP_bert, pixel_bert, cbt, MERLOT, lxmert} using terabyte-scale datasets crawled from multiple internet sources. When combined with large language models such as GPT-2 \cite{GPT2}, BERT\cite{bert}, Sentence-BERT\cite{sentence_bert}, and the T5 variants \cite{T5, flan_t5}, these vision-language models can be applied to several downstream tasks such as VQA \cite{antol2015vqa} and image captioning~\cite{xu2015show,yao2017boosting}.
Similar to other vision-language models that are trained with large language input, our approach leverages the state-of-the-art image captioning models~\cite{ofa_captioning} to provide a consistent description of past actions.

\section{Language-guided prediction}
\begin{figure*}[!ht]
    \centering
    \includegraphics[width=0.9\linewidth]{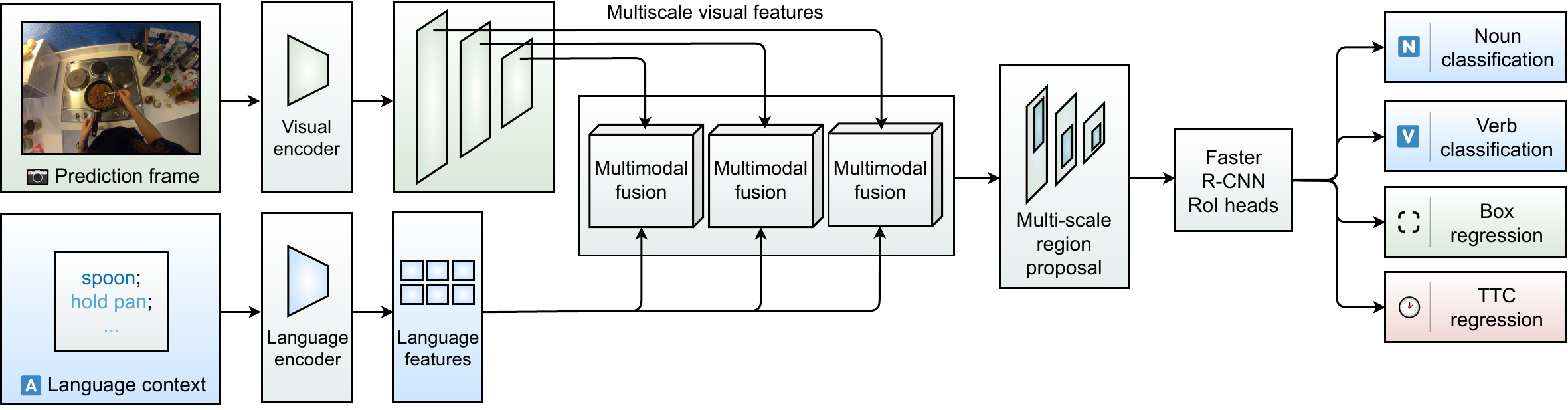}  
    \caption{\textbf{Overview of the \methodname model.} \methodname takes the prediction frame and the action context summary as input and predicts the bounding box of the next-active object, the noun-verb pair describing the associated action, and the time to contact (TTC). 
    Feature maps of different scales are extracted from the visual encoder and then fused via a multimodal fusion module with the encoded language features. Their output is then processed by a regular feature pyramid network (FPN), denoted as multi-scale region proposal networks, before being fed into the Faster R-CNN detector.
    \vspace{-0.4cm}
    }
\label{fig:architecture_overview}
\end{figure*}

We use language-based context summaries to represent the spatial-temporal context formed by past actions and object relationships. 
The context summary offers an explicit yet compact representation, which allows us to focus on task-relevant content and filter out unimportant information from the likely cluttered environment. To this end, we propose \methodname, a transformer-based model that leverages a joint-attention fusion mechanism, to effectively learn from the past action context to predict object interactions in the future.   
In the following, we describe our framework that takes video clips as input and anticipates the next object interaction. 
We begin by defining the object interaction anticipation task (Sec.~\ref{sec:task-def}). We then show how to generate a summary from an egocentric video to represent the action context provided in the past frames (Sec.~\ref{sec:caption generation}). Finally, we present our multimodal fusion model \methodname (Sec.~\ref{sec:laimformer}).   

\subsection{Task definition}
\label{sec:task-def}
Given a video $\mathcal{V}=\{f_1, \dotsm, f_T\}$ of length $T$ as input, our goal is to predict the next object interaction $\mathcal{O}$, in the next video frame $f_T$. We call $f_T$ the \textit{prediction frame}. 

As illustrated in \autoref{fig:Teaser.drawio}, the task is to anticipate object interaction in 
$f_T$.
$\mathcal{O}_T=\{o_1^T, \dotsm, o_N^T\}$ denotes the set of future object interactions, with $N>1$ whenever multiple objects are interacted with simultaneously. In other words, multiple $o_i^T$ can be detected in the prediction frame $f_T$.
For example, $N=2$ when one hand is holding two mugs in a kitchen scene, or the left hand is touching a plant while the right hand is moving a pot in a garden scene. Each object interaction $o_i^T = (b_i^T, n_i^T, v_i^T, t_i^T)$ consists of the object bounding box $b_i^T$, the semantic object label (noun) $n_i^T$, the action label (verb) $v_i^T$, and the time to contact (TTC) $t_i^T$. 
The task is to predict the next active object location $b_i^T$ in the prediction frame $f_T$, the associated action described by the verb-noun pair $(n_i^T, v_i^T)$, and the time $t_i^T$ to the point of contact when the interaction is going to start.
The task is set up such that the actual object interaction takes place after a buffer time $\Delta$ on frame $f_{T+\Delta}$ and the predicted time to contact $t_i^T=\Delta$. 
Ego4D uses $\Delta \geq 0.033 s$. This requires anticipating future actions and increases the task difficulty. 

\subsection{Summarizing the past}\label{sec:caption generation}

Anticipating object interactions in the future requires an understanding of what happened in the past. We first infer the per-frame \textit{action context} from each egocentric video. This process consists of extracting 
\textit{action descriptions}, \textit{held objects} and \textit{salient objects}. These information sources are aggregated independently across frames and finally forwarded as the action context %
to our anticipation model.

\myparagraph{Extracting frame-wise \textit{action descriptions}}
 Inspired by the label format used in egocentric video datasets~\cite{ego4d,damen2022rescaling}, we aim to describe the past actions by sequences of verb-noun pairs (e.g. ``wash tomato", ``cut tomato"), so-called \textit{action descriptions}. 
For each video frame $f$, we extract the action currently performed by the agent, represented by a single pair $(v, n)$ of a verb $v$ and a noun $n$.

We use an off-the-shelf image captioning model~\cite{ofa_captioning} to generate \textit{multiple} %
full-sentence captions for each frame of interest, so as to find repeatedly occurring outputs and reduce noise:
we perform part-of-speech tagging~\cite{akbik2018contextual} on these captions and collect \textit{candidate pairs} consisting of verbs followed by nouns within some cutoff distance $d \in \mathbb{N}_0^+$. 
Setting $d > 0$ allows us to account for additional words appearing between a verb-noun pair 
(e.g. ``\textit{eat apple}" from ``A person \textit{eating} \underline{a} \underline{red} \textit{apple}") and thereby obtain more valid candidate pairs from the captions. 
However, larger cutoff distances may introduce more spurious detections (e.g.\ ``\textit{eat gathering}", from ``A person \textit{eating} \underline{while} \underline{at} \underline{a} \textit{gathering}"). We use $d=4$ in our experiments.
The most frequent candidate pair across the frame's captions is selected as the verb-noun action description of this frame. 
If no valid verb-noun pair is found, the frame will have no action description.

\myparagraph{Extracting frame-wise \textit{held objects}}
As humans use their hands to interact with most objects, the sequence of objects held by the actor may be indicative of the overall task and can help the model better infer the immediate next steps. 
Motivated by this, we use an existing hand-object interaction predictor~\cite{VISOR2022} jointly with an object detector~\cite{unidet} to look for held objects and their semantic labels in each frame. 
Usually, such objects occur only in a subset of frames (e.g. due to the visibility of hands).

\myparagraph{Extracting frame-wise \textit{salient objects}}
Another characteristic shared by many daily action sequences is that the next active object is likely to have already appeared in the actor's view before the upcoming interaction. The current environment %
in the recording may also provide model conditioning. Therefore, we select a set of \textit{salient objects} in each frame as (1) the potential candidates for future active objects and (2) a proxy of the surroundings. 

We use CLIP \cite{CLIP}, %
a pre-trained open-vocabulary zero-shot classifier, %
to compute the cosine similarities between the CLIP image embedding of a frame and the embeddings of the object categories from the Ego4D domain. 
We choose to keep the highest-scoring $k$ objects as the set of salient objects for a given frame. 
Note that a small $k$ may fail to provide sufficient information, while a large $k$ makes capturing the true salient objects more likely, %
yet introduces more noise due to false detections. We set $k$=5. %

\myparagraph{Aggregating summaries over multiple frames}
After the frame-wise summarization step, we have, 
for each frame $t$, at most one verb-noun pair describing the action $a_t=(n, v) \in \mathcal{A}^{F}_t$, a set of hand-held objects $\mathcal{N}^{F}_{h,t}$, %
and a set of salient objects $\mathcal{N}^{F}_{s,t}$.
We then apply a \textit{cross-frame aggregation} scheme to each of the three context representations 
$\{ \mathcal{A}^{F}_t, \mathcal{N}^{F}_{h,t}, \mathcal{N}^{F}_{s,t} \}$, 
so as to identify contiguous segments of frames exhibiting the same activity, held or salient objects, while accounting for individual noisy frames possibly not matching the frame-wise summaries of their neighbors. 
The aggregation processes the frames in temporal order. 
We consider frames belonging to the same atomic action as an action segment, which is initiated by observing a number of reoccurrences of some verb-noun pair and terminated by a lack of its occurrence within a number of contiguous frames. After the aggregation, we obtain $\mathcal{A}$, the sequence of all action segments in the video.
We conduct identical aggregation processes to obtain $\mathcal{N}_{h}$ and $\mathcal{N}_{s}$. For each prediction frame, we construct its action context from a number of action segments preceding the frame, and the corresponding objects from $\mathcal{N}_{h/s}$. 

By aggregating across frames, we thus capture dominant actions and objects in the past, reduce noise in the generated action context and represent long timespans of similar activity by short textual descriptions.
The complete action context %
is provided to our model by concatenating the textual representations of the aggregation results of any desired subset of the three context representations $\{\mathcal{A}, \mathcal{N}_h, \mathcal{N}_s\}$, where we drop the index $t$ for convenience.
More details %
can be found in the Supp. Mat.

\subsection{\methodname: Multimodal fusion}
\label{sec:laimformer}
\autoref{fig:architecture_overview} visualizes the architecture of the proposed multimodal \methodname model. \methodname employs two different base encoders, for language and image input respectively. 
The input prediction frame $f_T$ is processed by a regular CNN-based visual encoder~\cite{resnet},  
producing a set of multiscale feature maps $\mathcal{F}_\mathcal{V}^s (f_T)$, with $s$ indicating the scale. %
The frame's action context is encoded by a language encoder~\cite{sentence_bert} and yields $\mathcal{F}_\mathcal{L}(\mathcal{C}_L)$. 
The feature maps belonging to different scale levels from the visual encoder are fused with the encoded language features via a joint-attention scheme. The result is then processed by a regular feature pyramid network (FPN)~\cite{fpn} before being forwarded to the Faster R-CNN~\cite{faster_rcnn} detector. 

\begin{figure}[t]
    \centering
    \includegraphics[width=0.8\linewidth]{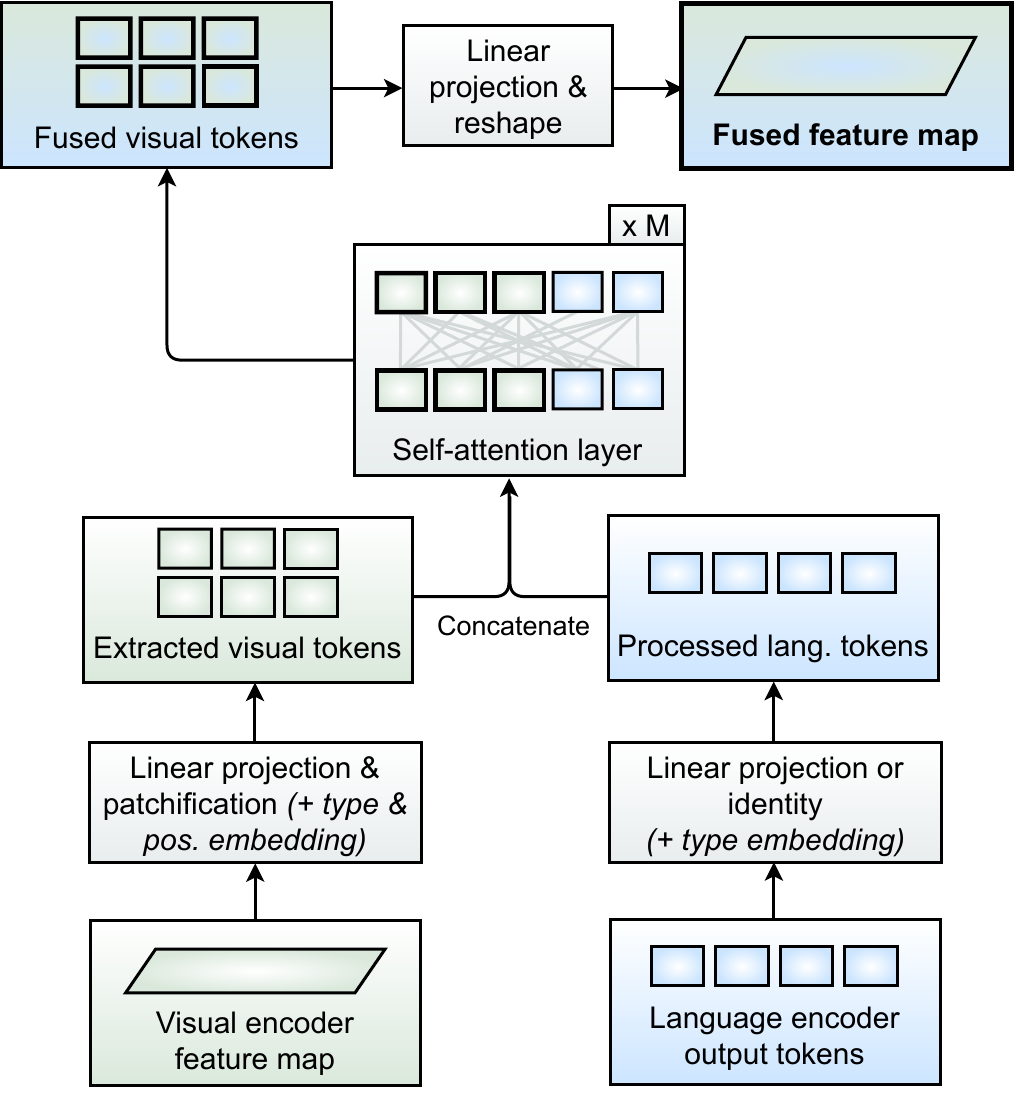}
    \caption{\textbf{Multimodal fusion module.} We first project the CNN feature map and the language tokens to a common dimensionality before adding the specific embeddings. We concatenate the visual and language tokens and feed them to $M$ self-attention layers. At the output, the fused visual tokens are projected back into the initial feature map shape.
    }
    \label{fig:fusion_module}
    \vspace{-0.5cm}
\end{figure}

\myparagraph{Multimodal fusion}
For simplicity, we describe the multimodal fusion module for a single input scale $s$, which we hence omit from notation. See~\autoref{fig:fusion_module} for a detailed view of a single Transformer Encoder layer together with the input projection stages. 
For the visual modality, we first split the image into a sequence of tokens. 
Specifically, the visual features $\mathcal{F}_\mathcal{V} (f_T) \in \mathbb{R}^{C \times H \times W}$ are reshaped to $\mathbb{R}^{N \times (P^2 \cdot C)}$, where $P$ is the token size and $N = \frac{H \times W}{P^2}$ the number of tokens. 
We add unique type embeddings as well as a positional encoding to each modality. 
Both visual and language tokens are first mapped to $\mathbb{R}^D$ via a linear projection before being fed to the fusion module in a concatenated form.
The fusion block consists of a regular Transformer encoder layer replicated $M$ times.
Finally, the visual tokens are regrouped into their initial shape to compose the feature map for the feature pyramid network (FPN).

\myparagraph{Multiscale fusion} 
The multimodal fusion block outputs multiple feature maps of varying scales: $\mathcal{F}_{fi} \in \mathbb{R}^{C_i \times H_i \times W_i}$, where $i$ is the scale index. They are fed into FPN, which has different parameters for each feature map, allowing the network to learn specialized features for each downsampling ratio.  
The output of FPN is processed by a Faster R-CNN detector. 

\myparagraph{Prediction head networks}
The Faster R-CNN detector outputs the bounding box $b$ through a regression layer, an associated object $n$ and verb $v$, and time to contact $t$.

\myparagraph{Training objective}
We train \methodname using the standard Faster R-CNN \cite{faster_rcnn} 
objective $\mathcal{L}_{box}$ for bounding box prediction, which consists of the localization and objectness, cross-entropy losses $\mathcal{L}_{noun}$ and $\mathcal{L}_{verb}$ for noun and verb prediction respectively, and an L1 loss $\mathcal{L}_{ttc}$ for time to contact prediction. 
The overall objective is 
\begin{equation}
\mathcal{L} = \mathcal{L}_{box} + \mathcal{L}_{noun} + \mathcal{L}_{verb} + \mathcal{L}_{ttc}.     
\end{equation}

As defined in Faster R-CNN, $\mathcal{L}_{box}$ is 
\begin{equation}
    \mathcal{L}_{box} = \frac{1}{N_\text{cls}} \sum_i \mathcal{L}_\text{cls} (p_i, p^*_i) + \frac{\lambda}{N_\text{reg}} \sum_i p^*_i \cdot L_\text{reg}(b_i - b^*_i),
\end{equation}
where $p_i$ is the predicted probability of a box being classified as a foreground object (true positive), $p_i^*$ is the ground-truth label; $b_i$ and $b^*_i$ are the predicted and ground-truth box coordinates. $N_\text{reg}$ is the number of bounding boxes used in training, $N_\text{cls}$ the detection network's image batch size. We leave $\lambda=11$ as in the reference implementation.

\section{Experiments}

\myparagraph{Dataset}
We use the dataset provided for the short-term object interaction anticipation task in \textbf{Ego4D}~\cite{ego4d}.
It contains various activities ranging from gardening work to cooking or car parts changing. 
There are in total 64,798 annotated video clips of 5 minutes each with frame rate $fps=30$. The dataset has highly imbalanced long tail distributions including 87 nouns and 74 verbs (e.g. the ``take'' class is almost 40\% of all the verb labels while ``mold'' appears only in one video). The train set consists of about 28k samples, while the rest are roughly equally split between validation and test . 
We also experimented with the EK100 dataset~\cite{damen2022rescaling} that consists of unscripted, egocentric kitchen activities together with interaction labels for 300 objects and 97 actions. 

\myparagraph{Evaluation}
As in Ego4D, we use the Top-5 mean Average Precision (mAP) to evaluate the performance of individual predictions. It considers two constraints: 
The \textbf{IoU} constraint requires that the predicted boxes are counted as hits only with intersection over union $\text{IoU} \geq 0.5$ with the ground-truth box; the \textbf{TTC} constraint requires that the predicted time-to-contact $t_i$ is close enough to the ground-truth $\hat{t_i}$, i.e. $|t_i - \hat{t_i} |< T_\delta, T_\delta=0.25$. 
\begin{itemize}[noitemsep,nolistsep]
    \item \textbf{Noun} refers to the Noun-Box Top-5 mAP. In addition to the IoU constraint, exact matches with the ground-truth nouns are required.
    \item \textbf{Noun-Verb} refers to the Noun-Verb-Box Top-5 mAP. Exact noun and verb matches with the IoU constraint.
    \item \textbf{Noun-TTC} refers to the Noun-Box Top-5 mAP. Exact noun matches with the TTC constraint.
    \item \textbf{Noun-Verb-TTC \textit{(``Overall")}} refers to the overall Top-5 mAP, intersecting exact noun-verb matches with IoU and TTC constraints.
\end{itemize}

Note that as detecting the next-active objects is the main focus of the task, all evaluation metrics are conditioned on the IoU constraint. 

\myparagraph{Implementation details}
We make use of the pre-trained Torchvision Faster R-CNN. 
ResNet-50 \cite{resnet} is used as the visual encoder and Sentence-BERT (SBERT)~\cite{sentence_bert} is used as the language encoder. 
We freeze ResNet-50 and apply the multimodal fusion scheme on top. 
We use the RAdam \cite{radam} optimizer with a weight decay of $2e^{-4}$, and an effective batch size of 32.
Besides the default Faster R-CNN background class in nouns, we also add it in the verb class which improves the Noun-Verb mAP by almost 2 points. We reduce the region proposal network's sampling batch size and the detection network's image batch size 
and use %
data augmentation techniques to reduce overfitting.
The number of verb-noun pairs used in the input is defined as the context length $L_c$. 
On average, an action segment described by one verb-noun phrase corresponds to one second with 30 frames, computed from the GT annotations. 
Unless specified otherwise, we use $L_c$=3.

\begin{table}[t!]
    \centering
    \small
    \begin{tabular}{c|c|c|c|c|c}
    \hline
    Set & Model &  N $\uparrow$ & N-V $\uparrow$ & N-T $\uparrow$ & A $\uparrow$\\
    \hline
    \multirow{3}{*}{Val} & Ego4D~\cite{ego4d} &  17.55 & 5.19 & 5.37 & 2.07 \\ 
   & TF (ours) & \textbf{20.19} & \textbf{7.55} & \textbf{6.13} & \textbf{2.60} \\
    
    \cdashline{2-6}
    & Improvement & 15.3\% &  45.4\% & 10.6\% & 25.6\% \\
    \hline
    \multirow{3}{*}{Test}  & Ego4D~\cite{ego4d} &  20.45 & 6.78 & 6.17 & 2.45 \\ 
    & TF (ours) & \textbf{24.69} & \textbf{9.97} & \textbf{7.33} &
    \textbf{3.44}\\
    \cdashline{2-6}
    & Improvement & 25.6\% & 47\% & 18.8\% & 40.4\% \\
    \hline
    \end{tabular}
    \vspace{-0.2cm}
    \caption{\textbf{Comparison with the state-of-the-art.} Our method \methodname (TF) outperform Ego4D by a large margin in all metrics. We report top-5 Noun (N), Noun-Verb (N-V), Noun-TTC (N-T), and Overall (A).%
    }
    \vspace{-0.1cm}
    \label{tab:sota}
\end{table}

\begin{figure}[!t]
    \centering
\includegraphics[width=0.75\linewidth,trim={1cm 0cm 2cm 0cm}]{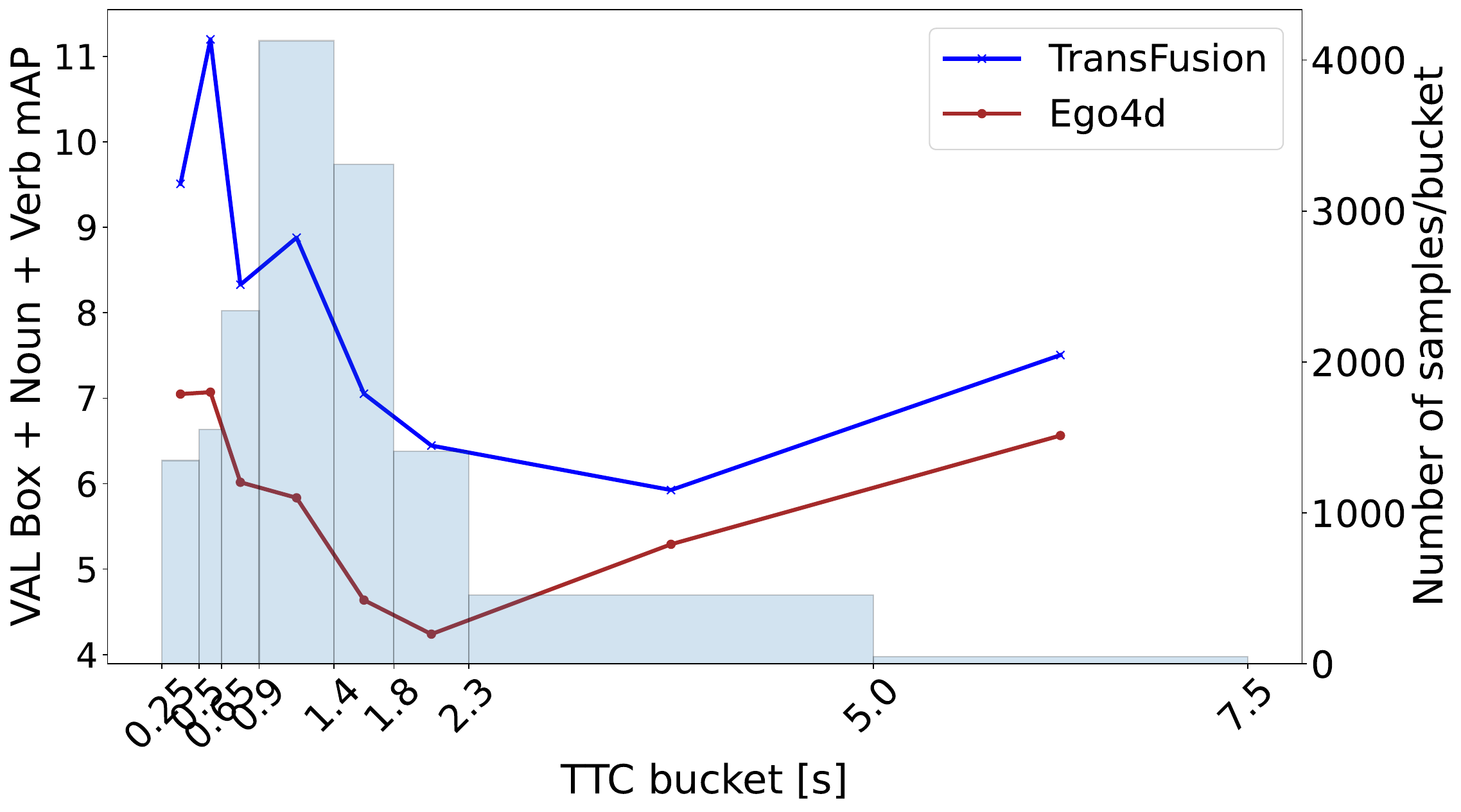}
    \vspace{-0.2cm}
    \caption{\textbf{Prediction of Noun-Verb in relation to time-to-contact.} Histogram of the time-to-contact labels is shown in blue bars. Performance measured in Noun-Verb is plotted as a function of time-to-contact.}
    \vspace{-0.6cm}
    \label{fig:longtail}
\end{figure}

\subsection{Comparison with state-of-the-art}
We first compare our method against the state-of-the-art Ego4D method~\cite{ego4d} on the validation and test sets. The \textbf{Ego4D} method employs a two-stage approach consisting of a ResNet-based Faster R-CNN detector and a SlowFast \cite{slowfast} 3D CNN video processing module. The Faster R-CNN outputs bounding boxes and noun predictions on the prediction frame, without using any video features. The detected boxes are used to perform region of interest (ROI) pooling \cite{Fast_RCNN} on the corresponding SlowFast 3D CNN video features. The pooled video features predict the verb and the time-to-contact values.

\begin{figure}[t]
    \centering    \includegraphics[width=0.75\linewidth,trim = {1.8cm 0cm 1.8cm 1.2cm}]{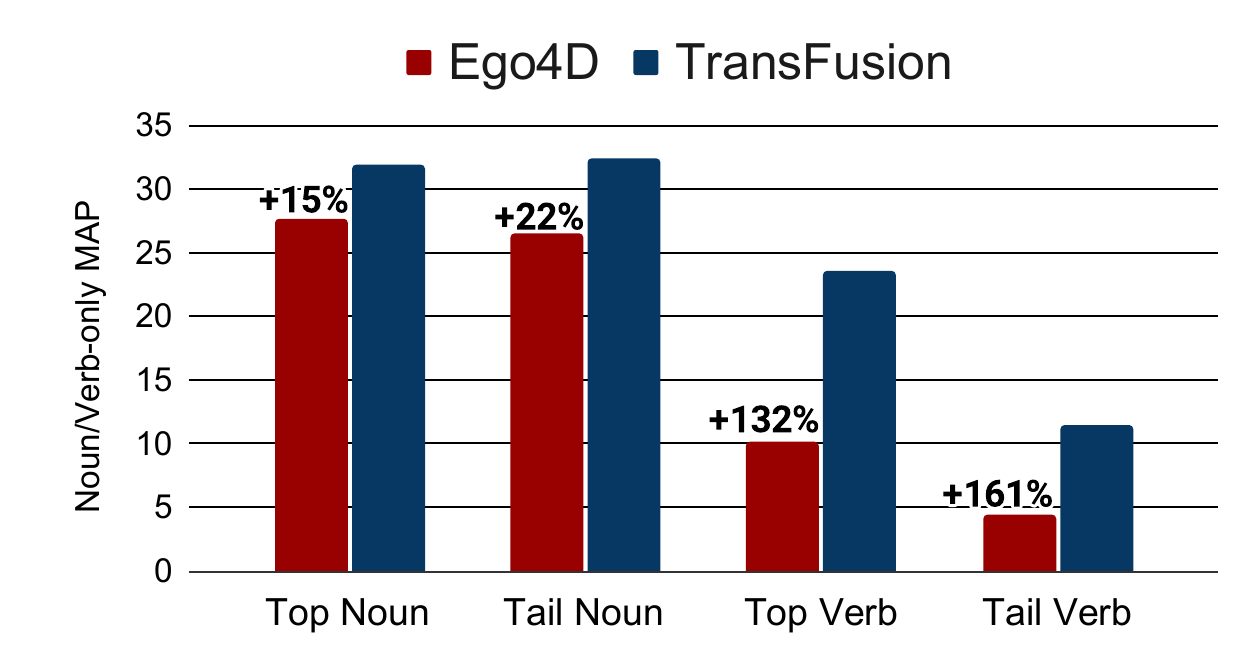}
    \vspace{-0.4cm}
    \caption{\textbf{Classification performance on top/tail categories.} We show the relative and absolute gains of \methodname over Ego4D for Noun-Only and Verb-Only mAP (without the IOU constraint). Relative improvements are written on top of the red bars. 
    }
    \vspace{-0.1cm}
\label{fig:classif_histo}
\end{figure}

\autoref{tab:sota} shows that \methodname significantly outperforms the state-of-the-art. 
This demonstrates the advantages of modeling action context explicitly for the interaction anticipation task.
\autoref{fig:longtail} shows the long tail distribution of the time-to-contact labels in the validation set. We also see that \methodname consistently outperforms Ego4D on the N-V metric. %
From the results shown in \autoref{fig:classif_histo}, we see that our method not only outperforms the Ego4D method on top noun and verb classes but also achieves bigger gains in the tail classes, demonstrating the generalization ability of using language-based context summaries. Top and tail buckets represent roughly 50\% of the samples, corresponding to 10 noun classes and 5 verb classes. 
 
\begin{figure}[t]
 \hspace{0.4cm} GT: open drawer 0.1s \hspace{1cm} GT: take phone 1s 

    \centering
        
     \begin{subfigure}[t]{.49\linewidth}
     \centering
    \includegraphics[width=.95\textwidth]{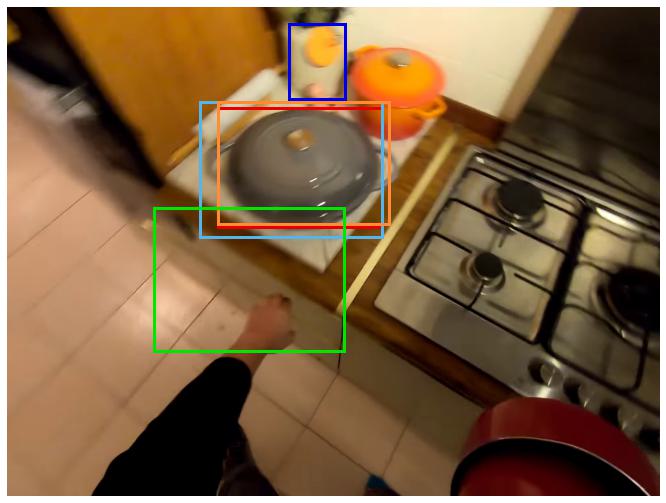}
    \vspace{-0.2cm}
    \caption{
    \small{
    Ego4D:  
    \textcolor{blue}{take container 0.78s},
    \textcolor{orange}{take container 0.46s},
    \textcolor{cyan}{take pot 0.42s},
    \textcolor{red}{take lid 0.46s}}
    }
    \end{subfigure}   
     \begin{subfigure}[t]{.49\linewidth}    
     \centering
     \includegraphics[width=.95\textwidth]{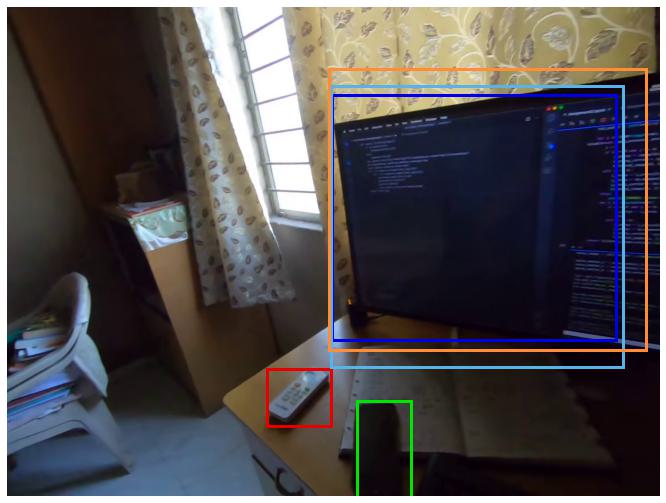} 
    \vspace{-0.2cm}     
    \caption{ 
    \small{
    Ego4D: 
    \textcolor{blue}{take table 0.78s}, 
    \textcolor{orange}{take container 0.85s},
    \textcolor{cyan}{take computer 0.86s},
    \textcolor{red}{take phone 1.11s}}
    }
    \end{subfigure}

     \begin{subfigure}[t]{.49\linewidth}
     \centering
    \includegraphics[width=.95\textwidth]{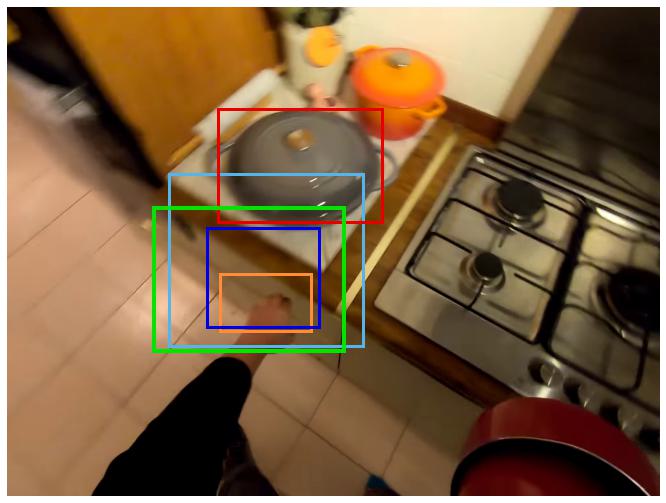}
        \vspace{-0.2cm}
    \caption{
        \small{
    TF (ours):
    \textcolor{blue}{open drawer 0.56s},
    \textcolor{orange}{open drawer 0.36s},
    \textcolor{cyan}{open drawer 0.44s},
    \textcolor{red}{take lid 0.6s}}}
    \end{subfigure}   
     \begin{subfigure}[t]{.49\linewidth}    
     \centering
     \includegraphics[width=.95\textwidth]{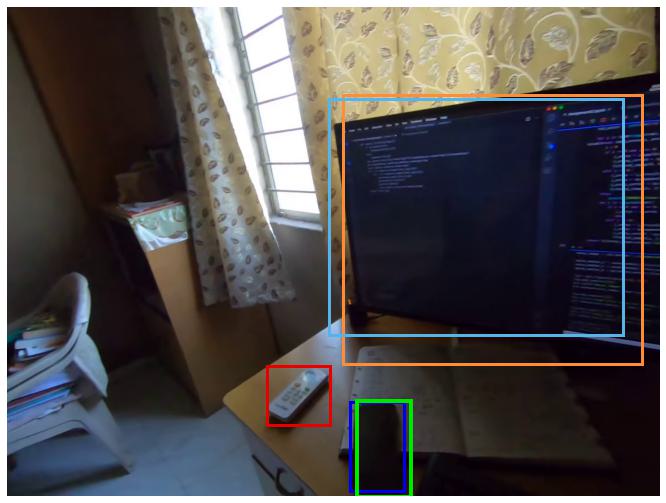} 
     \vspace{-0.2cm}
    \caption{ 
        \small{
    TF (ours):
    \textcolor{blue}{take phone 0.87s},
    \textcolor{orange}{open computer 0.63s},
    \textcolor{cyan}{operate tablet 0.73s},
    \textcolor{red}{take phone 0.98s}}}
    \end{subfigure}
    \vspace{-0.2cm}
    \caption{\textbf{Qualitative results of the proposed approach.} Ego4D results are shown in the top row and TransFusion in the bottom row. Green denotes the ground truth bounding box and the labels are on top of each column. %
    }
    \vspace{-0.6cm}
    \label{fig:viz_comp}
\end{figure}

 \autoref{fig:viz_comp} shows that the Ego4D approach manages to detect the most salient objects in the surrounding, but fails to reliably associate the bounding boxes to the correct action phrase. In contrast, \methodname is able to correctly associate the drawer bounding box to "open drawer" and to predict the correct action associated with the phone. 
 More examples are provided in the Supp. Mat.

\subsection{Evaluation of action context representation}
 
In \autoref{tab:annotation}, we ablate different variants of context representation and compare
against the representation that uses ground-truth labels (GT) provided by Ego4D, to assess the quality of our generated context summaries. 
The observed TTC Average Precision difference was around 2\%, thus omitted from the comparison. 
We report on the Noun and Noun-Verb metrics as we expect major differences in semantic-related tasks.

\begin{table}[t]
    \centering
    \small
    \setlength\tabcolsep{3.5pt} %

    \begin{tabular}{c|c|c|c|c|c|c||c}
    \hline
     Rep. & $\mathcal{A}$ & $\mathcal{N}_h$ & $\mathcal{N}_s$ &  $\mathcal{A}$+$\mathcal{N}_h$ & $\mathcal{A}$+$\mathcal{N}_s$ & All & GT\\
    \hline
    
    N $\uparrow$ &20.09 & 19.33 & 19.75 &  19.03 & 
    20.19 & \textbf{20.73} & 21.71\\
    N-V $\uparrow$ & 7.16 & 6.86 & 7.05 & 7.12 & 
    \textbf{7.55} & 7.32 %
    & 7.86 \\
    \hline
    \end{tabular}
    \vspace{-0.2cm}
    \caption{\textbf{Comparison of different action context representations.} We evaluate different variations of context representations on the predicted noun (N) and noun-verb (N-V), and compare them against ground-truth annotations (GT) provided by the Ego4D dataset. GT only contains action descriptions of verb-noun pairs. Action context represented by $\mathcal{A}$+$\mathcal{N}_s$ achieves the best N-V performance.   
    \vspace{-0.4cm}
}    
\label{tab:annotation}
\end{table}

Experiments %
show that we achieve results comparable to GT, and better than the Ego4D results shown in~\autoref{tab:sota}, demonstrating the effectiveness of our context representations. 
Action descriptions $\mathcal{A}$ consistently boost the verb classification performance.
Adding salient objects $\mathcal{N}_s$ to $\mathcal{A}$ further improves task performance. 
The effectiveness of $\mathcal{N}_h$ is limited by the accuracy of hand-object detection (see Supp. Mat. for details), but it does help to improve the prediction of nouns comparing the results of All and $\mathcal{A}$+$\mathcal{N}_s$.
These together offer a reasonable picture of using language summaries to represent the action context. In addition, we notice a domain gap in the choice of words between the generated labels and the ground-truth ones. In some cases, even if the description is satisfactory for a human evaluator, the information may be too vague to be directly useful to the model. Phrasing differences between the generated captions and the training corpus of the language encoders may further help to improve the performance. 
We use $\mathcal{A}$+$\mathcal{N}_s$ to represent the action context, as a balance between computational cost and accuracy. 
Ablations on the hyperparameters $d$, the cutoff distance, and $k$, the number of salient objects per frame, show that the performance is less sensitive to their settings. More details can be found in the Supp. Mat.

\subsection{Ablation studies}

\myparagraph{Length of action context}
A significant parameter in obtaining good performance is the length of action context, i.e., how many past actions to include in the input. 
An adequate context length is important to allow the model to pick up more specific activity patterns.
Some activities that have a more structured nature or certain repeating patterns (e.g. cooking or repairing a bicycle) are easier to model with an increased context length compared to more random activities such as playing basketball. 
\autoref{fig:gt_labels_vs_slowfast} shows that the overall performance plateaus at $L_c=2$ and higher.

\begin{table}[t]
    \centering
    \small
    \begin{tabular}{c|c|c|c}
    \hline
    Language Encoder & Token size  & N $\uparrow$& N-V $\uparrow$\\
    \hline
    SBERT/BERT~\cite{sentence_bert,bert} & 384 &  \textbf{20.19} & 7.55 \\ %
    SBERT/RoBERTa~\cite{sentence_bert,robert} & 768 &  19.78  & \textbf{7.78} \\ %
    SBERT/RoBERTa* & 768 &  18.16  & 7.21 \\
    GPT-2~\cite{GPT2} & 768 &  18.98  & 7.08 \\ %
    Flan-T5~\cite{flan_t5} & 1024 &  18.26  & 6.61  \\ %
    \hline
    \end{tabular}
    \caption{\textbf{Language encoder ablations.} We compare SBERT using BERT and RoBERTa  backbones, with GPT-2 and Flan-T5. Token size shows the token output size of each language encoder. They are projected to a dimension of 768 before fusion. %
    *Denotes the run without finetuning the last language encoder layer.
    }
    \label{tab:language encoders}
    \vspace{-0.1cm}
\end{table}

\myparagraph{Language encoder}
A language encoder is used to process the context summaries in \methodname. 
To understand the impact of different language encoders, we experiment with multiple LLMs including SBERT with a RoBERTa backbone~\cite{sentence_bert,robert}, GPT-2~\cite{GPT2}, and FLAN-T5~\cite{flan_t5} using the validation set.
\autoref{tab:language encoders} summarizes the performance of different language encoders, showing that the SBERT models perform best. We consider TTC less relevant here. 
We hypothesize that the difference comes from the training objectives of language models. SBERT is trained to map semantically similar sentences closer to each other in the embedding space. This is particularly useful when the model needs to understand descriptions with semantically close words, e.g. ``cut carrots'' and ``slice carrots''.
GPT-2 and FLAN-T5 are optimized using generative objectives
which makes them overly sensitive to the input word choice.
The run without finetuning (denoted by *) performs worse than the finetuned equivalent, %
showing that domain adaptation is required by the language encoders to adapt to the summarization syntax and activity-specific vocabulary.

\begin{table}[t]
    \centering
    \small
    \begin{tabular}{c|c|c|c|c}
    \hline
    Parameter sharing & \# Parameters & L2 & N $\uparrow$ & N-V 
    $\uparrow$\\
    \hline
    \xmark & 122 mln & 2e-4 & \textbf{20.19} & \textbf{7.55} \\
    \cmark & 171 mln & 3e-5 & 19.18 & 6.94 \\
    
    \hline
    \end{tabular}
    \vspace{-0.2cm}
    \caption{\textbf{Multiscale fusion parameter sharing.} %
    }
    \label{tab:parameter_sharing}
    \vspace{-0.2cm}
\end{table}
 
\begin{table}[t]
    \centering
    \small
    \begin{tabular}{c|c|c|c}
    \hline
    Forwarding strategy & Copy & Simple & Residual \\ 
    \hline
    N $\uparrow$  &  \textbf{20.19} & 18.97 & 18.38 \\
    N-V $\uparrow$ & \textbf{7.55} & 7.27  & 6.96 \\
    \hline
    \end{tabular}
        \vspace{-0.2cm}
    \caption{\textbf{Language feature forwarding strategies.} }
    \vspace{-0.5cm}
    \label{tab:feature_forwarding}
\end{table}

\myparagraph{Fusion module ablation}
We further ablate the \methodname design choices and investigate whether the model can benefit from feature reuse at the different scale levels. Firstly, we investigate the effect of sharing the Transformer encoder parameters over the multiscale fusion levels. 
The new setup leads to a decrease in performance %
as can be seen in \autoref{tab:parameter_sharing}. Secondly, we experiment with forwarding the fused language tokens to the latter stages, with and without residual connections, as an alternative way of sharing fused features. The base \methodname implementation simply copies the encoded language features at each fusion level (see \autoref{fig:fusion_module}). The results are presented in \autoref{tab:feature_forwarding} where we notice the superior results obtained with the ``copy'' strategy as opposed to reusing features from previous scale levels. 

\subsection{Comparison to video features} 
We evaluate whether language summary can better represent action context than video frames for the object-anticipation task. To do so, we implement \textbf{\methodname-Video}, which takes video features extracted from the Ego4D SlowFast\cite{slowfast} model as input and uses the identical fusion module as in \methodname.

\begin{figure}[t!]
    \vspace{-0.3cm}
        \centering
        \includegraphics[width=.75\linewidth, trim={1.5cm, 0.2cm, 0.2cm, 0cm}]{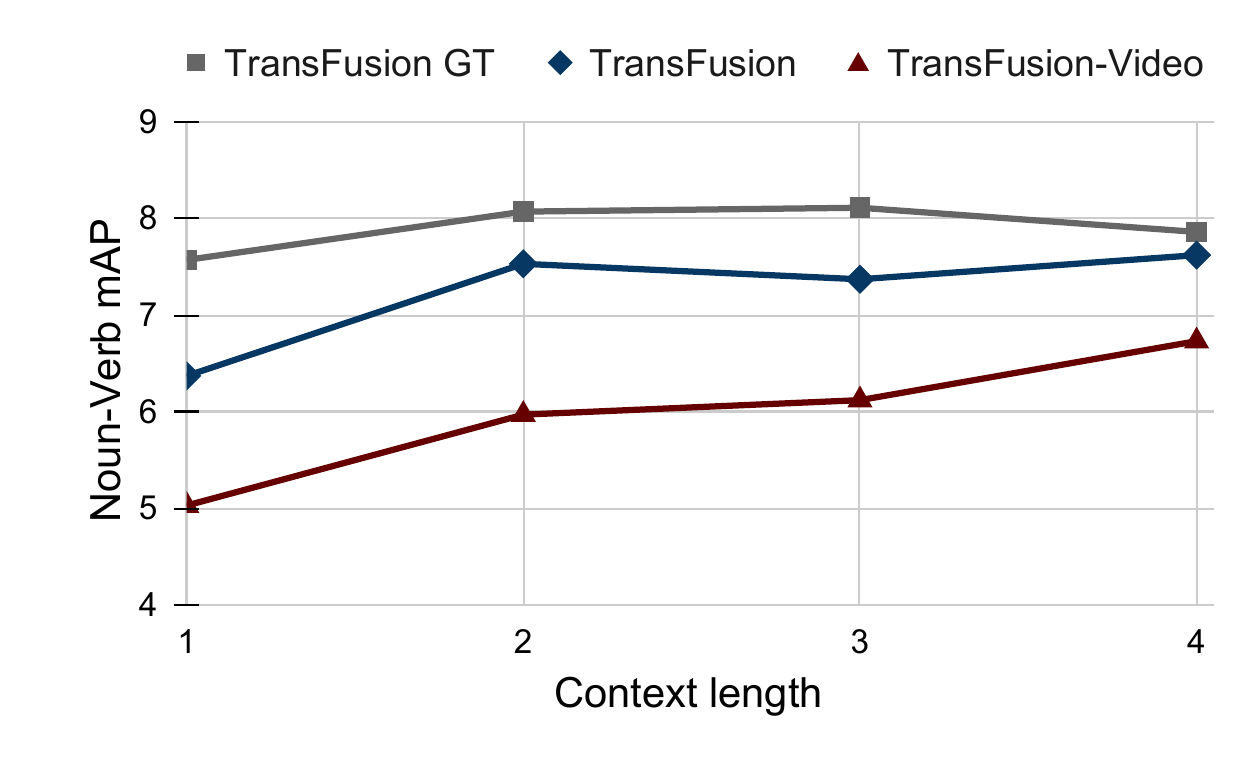}
        \vspace{-0.3cm}
    \caption{\textbf{Comparison to video features.} \methodname outperforms \methodname-Video over different context lengths measured by Noun-Verb mAP. \methodname using GT annotations represents the upper bound.  
    }
    \label{fig:gt_labels_vs_slowfast}
    
    \vspace{-0.4cm}
    
\end{figure}

\autoref{fig:gt_labels_vs_slowfast} shows that \methodname consistently outperform \methodname-Video. 
$L_c \times 30$ frames are taken in \methodname-Video. 
For video features, the domain gap between pre-train and target datasets seems too large to allow effective generalization to a diverse dataset such as Ego4D. We suspect that the feature representations are more entangled due to the need to represent temporal, visual, and semantic aspects simultaneously. With language, the temporal aspect is directly encoded in the succession of words. 

Given a video clip of one second, language can summarize it in two words whereas corresponding video features take up more than 12 times space (e.g. smaller SBERT~\cite{sentence_bert} feature of $2\times 384$ compared to SlowFast~\cite{slowfast} feature of $4\times2304$). 
Therefore, our language-based context representation has the advantage when it comes to longer video sequences.  
Computational costs measured in GFLOPs and latency for both \methodname and \methodname-Video models at both training and inference time are practically identical. Overall, \methodname obtains a 21\% increase in N-V performance for a similar computational budget.
See the Supp. Mat. for more details.

\subsection{Validating on EPIC-Kitchens 100}
To understand how our proposed model generalizes to other settings, we further test it on the EK100~\cite{damen2022rescaling} dataset. Since it consists of unscripted cooking activities performed by different actors, we can assess if our model is capable of picking up action regularities across different actors. Because the dataset only provides action annotations but not object-interaction anticipation labels, we preprocess the dataset annotations and videos using UniDet \cite{unidet}. 
We split the samples such that the model does not see the clips from the same video at train and validation, whereas in Ego4D, the model may encounter this. 
Notably, compared to Ego4D which has a single vegetable-fruit class, EK100 has a richer set of nouns including ``salad'', ``parsley'', ``bean'', ``tomato'' etc., and overall a more fine-grained domain. We train \methodname with Ego4D pre-trained Faster R-CNN on EK100 and compare against \methodname-Video.

\begin{table}[!t]
    \centering
    \small
    \begin{tabular}{c|c|c||c}
    \hline
    Model & TF & TF-Video & TF GT \\
    \hline
    N$\uparrow$ & \textbf{7.03} & 6.57 & 11.16  \\
    N-V$\uparrow$ & \textbf{3.80} & 3.72 & 5.65\\
    \hline
    \end{tabular}
    \caption{\textbf{Object interaction anticipation on EK100.}
    EK100 results using existing ground-truth (GT), generated summary, and video features. 
   TF stands for \methodname.} 
   \vspace{-0.3cm}
    \label{tab:ek100}
\end{table}

Results in \autoref{tab:ek100} 
suggest that \methodname learns to represent structured activities effectively with the inclusion of language features. 
EK100 videos have a lower resolution of 256$\times$456, in contrast to 1080$\times$1920 in Ego4D. 
This has a negative impact on our generated context summaries, and the overall gains of \methodname are more on par with TF-Video. The results using GT annotations indicate that better performance can be unlocked using more accurate context summaries. %
More details can be found in Supp. Mat.

\section{Conclusion and discussion}

We propose a multimodal fusion model, \methodname, that anticipates object interactions by considering past action context represented by language summaries.
Our experiments on two challenging egocentric video datasets demonstrate how language summaries can improve object interaction anticipation, highlighting the representational power of language descriptions.
While we do not consider any temporal information, it is possible to integrate motion cues like optical flow to improve the prediction of time to contact. 
We would like to highlight that the proposed approach is not limited to object interaction anticipation.
In general, language provides a universal interface to complement the visual input with information encoded in language models or task-specific pipelines. 
Future work can leverage language summary for other video reasoning tasks \cite{zeng2017leveraging,zhang2016video}, or extend it to describe the possible future activities \cite{lang2pose,kuo2022trajectory} for long-term interaction anticipation.

\paragraph{Acknowledgements} This work was supported by an ETH Zurich Postdoctoral Fellowship. We thank Adrian Spurr for the insightful comments and Velko Vechev for the voice-over.

{\small
\bibliographystyle{ieee_fullname}
\bibliography{main}
}

\clearpage
\newpage
\title{Summarize the Past to Predict the Future: Natural Language Descriptions of Context Boost Multimodal Object Interaction Anticipation \\\large{[Supplementary Material]}}
\author{}
\maketitle
\tableofcontents

\setcounter{table}{6}
\setcounter{figure}{7}

\section{Past summarization}
\label{sup:preprocessing}

\subsection{Method details}
\paragraph{Extracting frame-wise action context}

\begin{figure*}[!ht]
    \centering
    \includegraphics[width=1.0\linewidth]{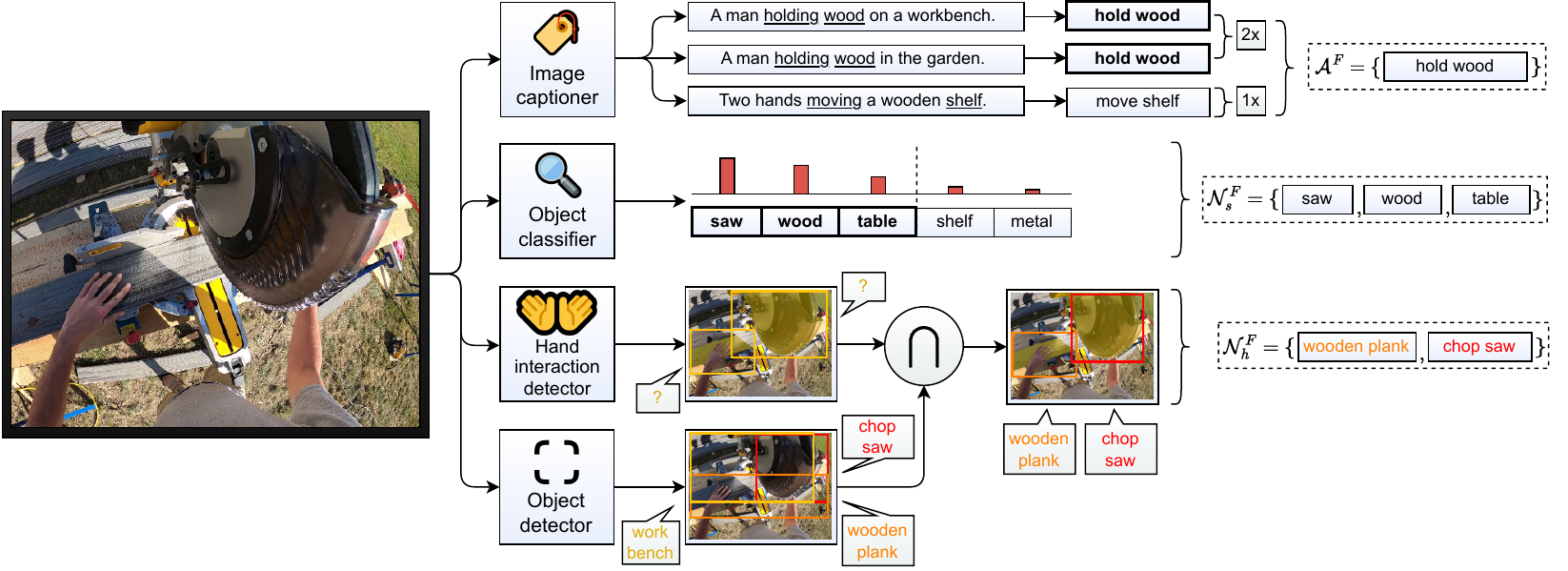}  
    \caption{\textbf{Illustration of the frame-wise context extraction.} Frame-wise context is extracted using off-the-shelf models: an image captioner for obtaining verb-noun pairs for $\mathcal{A}$, an object classifier to keep the highest-ranking $k$ objects (here $k=3$) for $\mathcal{N}_s$, and a hand-object interaction detector used jointly with an object detector to obtain and subsequently label active object bounding boxes for $\mathcal{N}_h$. Details can be found in Section 3.2 of the main paper and Section 1.1 of this supplement. 
    \vspace{-0.4cm}
    }
\label{fig:frame_wise_context_extraction}
\end{figure*}

A visualization of the frame-wise action context extraction is provided in \autoref{fig:frame_wise_context_extraction}.

To obtain action descriptions consisting of verb-noun pairs, we first generate multiple image captions (e.g.\ \textit{``a person cutting wood"}) by forwarding diverse prompts to the task-agnostic and modality-agnostic OFA model~\cite{ofa_captioning}. The exact prompts used are \textit{``what does the image describe?", ``what is the person in this picture doing?",} and \textit{``what is happening in this picture?"}.

We then perform part-of-speech tagging on the natural-language captions using Flair \cite{akbik2018contextual}, followed by a lemmatization using NLTK \cite{NLTK}, to extract candidate verb-noun pairs intended to represent the frame's action description (e.g. \textit{``cut wood"}). We obtain at most one verb-noun pair per processed frame by selecting the most frequently found pair. In the case of ties, we select the pair that was detected first.

The extraction of frame-wise salient objects $\mathcal{N}_s^F$ is described in Section 3.2 of the main paper.

To extract frame-wise held objects $\mathcal{N}_h^F$, we first obtain \textit{labelless} bounding boxes of active objects from EPIC-KITCHENS VISOR \cite{VISOR2022} together with labeled object bounding boxes (of not necessarily active objects) from \cite{unidet}, as further visualized in \autoref{fig:frame_wise_context_extraction}. To obtain \textit{labeled} bounding boxes of active objects, a pair of bounding boxes detected by UniDet and VISOR is considered to show the same object if they exhibit an intersection over union (IoU) greater than a threshold $\theta_{IoU}$, where we set $\theta_{IoU} = 0.25$. The labels corresponding to these bounding boxes together form the set $\mathcal{N}_h^F$ of held objects for this frame.

As we use UniDet, pre-trained on COCO \cite{coco}, for object detection in an off-the-shelf manner without further training, the domain of the object detector is not aligned with that of the Ego4D nouns. We thus perform some label merging to simplify the UniDet detection domain, e.g. merging ``home appliance" and ``pressure cooker" into ``machine".

\paragraph{Cross-frame aggregation}

\begin{figure*}[!ht]
    \centering
    \includegraphics[width=1.0\linewidth]{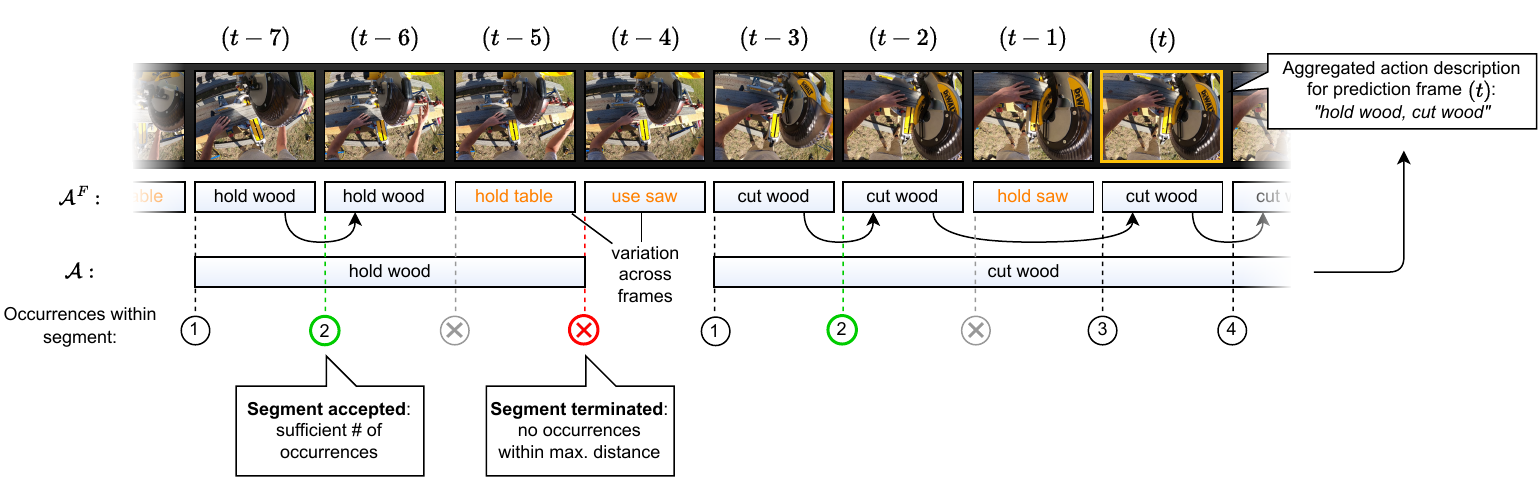}  
    \caption{\textbf{Illustration of the \textit{cross-frame aggregation} scheme, as used to construct action descriptions $\mathcal{A}$ from $\mathcal{A}^F$.} Identical schemes are used to construct $\mathcal{N}_s$ and $\mathcal{N}_h$. A sequence of identical frame-wise action descriptions (see \autoref{fig:frame_wise_context_extraction}) with the term $v$ forms a segment. As the aggregation traverses the frames, the segment is accepted into $\mathcal{A}$ once a number $P_{o,\mathcal{A}}$ of occurrences of $v$ have been found with each at most $P_{\ell,\mathcal{A}}$ frames apart from the last, and terminated once no occurrences have been found within $P_{\ell,\mathcal{A}}$ frames. The preceding $L_c$ segments up to a frame $(t)$ in $\mathcal{A}$ form the action description for that frame.
    \vspace{-0.4cm}
    }
\label{fig:cross_frame_aggregation}
\end{figure*}

See \autoref{fig:cross_frame_aggregation} for an illustration of the cross-frame aggregation scheme, restricted to $\mathcal{A}$ for simplicity. An identical scheme is used to extract $\mathcal{N}_s$ and $\mathcal{N}_h$.

For a given frame on which to predict, we mark the $150$ (Ego4D) resp.\ $120$ (EPIC-KITCHENS) previous frames to be processed by the context extraction models using a stride of $3$ frames. The videos in the Ego4D dataset use $30$ FPS, while for EPIC-KITCHENS we subsample the videos to 24 FPS. We thus process the preceding $5$ seconds for each prediction frame in both datasets. Yet, it is possible for a prediction frame to make use of action context obtained from more than $5$ seconds in the past through the inclusion of action context computed for previous prediction frames.

After processing the individual frames, we separately post-process each action context category $c \in \{\mathcal{A}, \mathcal{N}_h, \mathcal{N}_s\}$ (noun-verb pairs, held objects, and salient objects) via a cross-frame aggregation scheme to merge consecutive frames with identical terms into segments. Note how working with language summaries allows us to opt for this simple duplicate elimination scheme, whereas video or embedding-based input is often repetitive and not straightforward to deduplicate.

More specifically, let $\mathcal{V}_c$ represent the vocabulary of category $c$, as detailed in the next subsection. For instance, $V_{\mathcal{N}_s}$ is the domain of nouns in the Ego4D short-term object interaction anticipation dataset.

The aggregation progresses through the video in a temporal manner, maintaining vectors of active and past segments. Unless a segment with the given term $v \in \mathcal{V}_c$ is already active, a series of occurrences of $v$ in a sequence of frames that is at least $P_{o,c}$ long, with consecutive occurrences at most $P_{\ell,c}$ frames apart, lead to a segment being \textit{accepted} into the list of active segments. Note that the segment is considered to \textit{start} with the first occurrence that contributed to its acceptance, and is \textit{terminated} once $v$ has not occurred in the last $P_{\ell,c}$ frames. For a given context length $L_c$ to be used when constructing the action context for a prediction frame, we construct a context of at most $L_c$ non-overlapping active and/or past segments. Segments containing more occurrences of their respective term eliminate overlapping segments of different terms with fewer occurrences.

To construct the action context for a given prediction frame and action context category $c$, we distinguish between the \textit{current} context, and the \textit{past} context, with a context length of $L_c$ resulting in $1$ current and $L_{c}-1$ past segments for $\mathcal{A}$ and $\mathcal{N}_h$. For $\mathcal{N}_s$, we do not consider past segments. Instead, we operate only using currently active segments, as salient objects change quickly throughout video frames and we are interested in summarizing the \textit{recent} environment of the actor for this action context category.

We set $P_{o,\mathcal{A}} = 1$, $P_{o,\mathcal{N}_h} = 7$, $P_{o,\mathcal{N}_s} = 10$. We further use $P_{\ell,\mathcal{A}} = P_{\ell,\mathcal{N}_h} = P_{\ell,\mathcal{N}_s} = 7$.
For the experiments in Table 1, $L_\mathcal{A} = 3$. For those in Table 2, \autoref{tab:nao_gt_input} and \autoref{tab:gt_noun_occurrences}, $L_\mathcal{A} = 4$. In all cases, $L_{\mathcal{N}_h} = L_{\mathcal{N}_s} = 3$.

\paragraph{Action context vocabularies}

\begin{table}[t]
    \centering
    \small
    \begin{tabular}{c|c|c|c|c}
    \hline
    Model & Filtering & NO $\uparrow$ & N $\uparrow$ & N-V $\uparrow$\\
    \hline
    $\mathcal{A} + \mathcal{N}_s$ & \checkmark & \textbf{33.20} & \textbf{19.63} & \textbf{7.36} \\
    $\mathcal{A} + \mathcal{N}_s$ & \xmark & 32.38 & 19.27 & 7.20\\ 
    \hline
    $\mathcal{A}$ & \checkmark & \textbf{32.41} & \textbf{18.67} & \textbf{7.16} \\  
    $\mathcal{A}$ & \xmark & 30.82 & 18.05 & 6.61 \\
    \hline
    
    \end{tabular}
    \label{tab:filtered_ss_unfiltered_annots}    
    \caption{\textbf{Evaluation of filtering of extracted verb-noun action descriptions.} We evaluate the performance obtained on the Ego4D validation set when training using filtered verb-noun pairs with nouns restricted to the Ego4D noun domain, and when using unfiltered verb-noun pairs. The results show that the filtered versions achieve better performance scores than their unfiltered counterparts.}
    \label{tab:filtered_ss_unfiltered_annots}    
\end{table}

For the verb-noun action description pairs $\mathcal{A}$, we denote the vocabulary $V_\mathcal{A} = V_{\mathcal{A}, verb} \times V_{\mathcal{A}, noun}$. We restrict $V_{\mathcal{A}, noun}$ and $V_{\mathcal{N}_s}$ to the domain of the 87 noun classes used in the Ego4D dataset by eliminating all verb-noun pairs with nouns outside this domain during the cross-frame aggregation. To increase the number of frames for which action descriptions can be found, $V_{\mathcal{A}, noun}$ additionally contains a small set of generic words such as ``something" and ``object", as these occur frequently in the captions generated by OFA.

As seen in \autoref{tab:filtered_ss_unfiltered_annots}, restricting $V_{\mathcal{A}, noun}$ in this manner yields better performance for both $\mathcal{A}$ and $\mathcal{A} + \mathcal{N}_s$ models. We hypothesize that using a broad vocabulary might inhibit the model's ability to learn regularities in the language input, given the limited number of training samples available. $V_{\mathcal{N}_h}$ is the domain of UniDet object classes, while $V_{\mathcal{A}, verb}$ consists of the lemmatized versions of all verbs in the output domain of OFA.

\begin{table}[!t]
    \centering
    \begin{tabular}{c|c|c|c|c} 
        & \textbf{$\mathcal{A}$} & \textbf{$\mathcal{N}_s$} & $\mathcal{N}_h$ & SlowFast \\
        \midrule
       t (ms)  & 340 & 200 & 280 & 200 \\  
       M (GB)  & 5.5 & 3.0 & 5.4 & 9.5  
    \end{tabular}
    \caption{\textbf{Feature generation costs for a context unit}. We compare the time and the GPU memory requirements for generating each of the language inputs and the SlowFast features. Note that the generation of $A$, $\mathcal{N}_s$ and $\mathcal{N}_h$ can be parallelized, and we only utilize $A + \mathcal{N}_s$ for our final model.}
    \label{sup_tab:cc_summary_generation}
\end{table} 

\subsection{Computational cost and hyperparameters}

\paragraph{Computational cost}
The computational cost of generating language and video features is reported in \autoref{sup_tab:cc_summary_generation}.  
To obtain the language features, our final model configuration using $\mathcal{A}$ + $\mathcal{N}_s$ needs only about 140ms more time per frame than when using SlowFast features, while requiring less GPU memory: 8.5 instead of 9.5 GB. In this calculation, we consider the systems to generate the action context language features to be running in parallel. 
We consider the two methods to have similar costs.

\paragraph{Hyperparameter sensitivity}

We additionally present the effect that different context generation hyperparameters have on the quality of the generated action context. For $\mathcal{N}_s$, \autoref{sup_tab:k_sensitivity} shows the effect of $k$, the number of candidate salient objects that are kept per frame, using 3 metrics. Precision in \autoref{sup_tab:k_sensitivity} denotes the fraction of all inferred salient objects which are the respective frames' ground-truth NAO noun, while recall denotes how often the ground-truth noun appears in its frames' inferred salient objects. 
Similarly, \autoref{sup_tab:d_sensitivity} illustrates the effect of $d$, the maximum distance between verbs and nouns when extracting candidate verb-noun pairs from the natural-language image captions during $\mathcal{A}$ context construction. The number of exact hits in \autoref{sup_tab:d_sensitivity} represents how often the generated noun/verb matches the ground-truth Ego4D NAO noun/verb. The average GloVe \cite{glove} similarity in both tables is computed by averaging and then normalizing the 300-dimensional GloVe vector representation of the salient objects (for $\mathcal{N}_s$, \autoref{sup_tab:k_sensitivity}) resp. verbs/nouns (for $\mathcal{A}$, \autoref{sup_tab:d_sensitivity}) in the context description, and computing their dot product with the normalized GloVe embedding of the ground-truth noun/verb. It represents a less strict matching evaluation to account for the possibility of synonyms to the ground-truth: the closer the generated descriptions are, the higher the final average similarity is. Frame coverage shows how many frames we retain at least one salient object/verb-noun pair for after the cross-frame aggregation: larger numbers are better here since we reduce the risk of skipping important action steps. 
The average Glove similarity remains virtually the same for $k \geq 3$.

\paragraph{Sensitivity of $\mathcal{N}^F_h$ to noise from hand-object interaction and object detectors} To obtain a descriptive $\mathcal{N}^F_h$ for a given frame, both involved models, the hand-object interaction (HOI) detector and the object detector, must produce satisfactory results which can additionally be matched to each other. Specifically, the HOI detector must avoid omissions, false positives, undersegmentations and oversegmentations of active objects. The bounding boxes of active objects are obtained by taking the outer limits of the segmentation. The object detector must detect the active object and additionally assign it a correct label. Lastly, the bounding boxes produced by both models must sufficiently overlap so that the label inferred by the object detector can be assigned to the object segmented by the HOI detector. Examples of $\mathcal{N}^F_h$ we deem useful to the prediction task are visualized in \autoref{sup_fig:n_h_good}. We further showcase some failure cases of the HOI detector in \autoref{sup_fig:n_h_visor_fail} and of the object detector in \autoref{sup_fig:n_h_unidet_fail}. These failures lead to missing or incorrect $\mathcal{N}^F_h$.

\begin{table}[!t]
    \centering

    \begin{tabular}{c|c|c|c|c} 
        $k$ & Precision & Recall & Ø GloVe 
        sim. & Frame coverage \\ 
        \hline
        1 & \textbf{0.2582} & 0.2566 & \textbf{0.4221} & 57.69 \% \\
        2 & 0.1892 & 0.2715 & 0.3656 & 82.13 \% \\
        3 & 0.1550 & 0.3106 & 0.3371 & 89.82\% \\
        4 & 0.1370 & 0.3374 & 0.3198 & 92.80\% \\ 
        5 & 0.1275 & \textbf{0.3477} & 0.3113 & \textbf{93.93} \%  
    \end{tabular}

    \caption{\textbf{Influence of parameter $k$ on generated $\mathcal{N}_s$ action context}. Increasing $k$ leads to better frame coverage (fraction of frames for which $\mathcal{A}^F$ is non-empty) and increased recall of the generated $\mathcal{N}_s$ on the Ego4D validation set, but reduces their precision and GloVe similarity to the ground-truth next active object nouns. As we consider recall and frame coverage to be more important than precision, and noisy detections are likely to be eliminated by the subsequent cross-frame aggregation, we choose $k=5$.}
    \label{sup_tab:k_sensitivity}
\end{table}

\begin{table}[!t]
    \centering
    \begin{tabular}{c|c|c|c|c} 
        $d$ & Hits (N) & Hits (V) & Ø GloVe sim. * & Frame coverage \\
        \hline
        1 & 10.61\% & \textbf{3.42\%} & 0.3348 & 75.18 \% \\ 
        2 & 13.12\% & 3.29\% & 0.3567 & 78.85 \% \\
        3 & 15.69\% & 3.05\% & 0.3752 & 79.41 \% \\
        4 & \textbf{16.99\%} & 3.15\% & \textbf{0.3850} & 79.64 \% \\
        5 & 9.67\% & 2.65\% & 0.3426 & \textbf{81.11 \%} \\
    \end{tabular}
    \caption{\textbf{Influence of parameter $d$ on generated $\mathcal{A}$ action context}. We observe a sudden drop in the fraction of ground-truth correspondences for both nouns and verbs of $\mathcal{A}$ on the Ego4D validation set when transitioning from $d=4$ to $d=5$, likely caused by the introduction of spurious verb-noun pair detections. We hence choose $d=4$. *Average between cosine similarity of $\mathcal{A}$-noun to ground-truth noun and $\mathcal{A}$-verb to ground-truth verb, measured using GloVe embeddings.}
    \label{sup_tab:d_sensitivity}
\end{table}

\section{The TransFusion model}
\label{sup:model}

The fusion module is based on the query-key-value (QKV) attention mechanism popularized by the Transformer \cite{transformer} architecture. Such an attention aggregation scheme can loosely be
interpreted as computing a weighted average of the value vectors \textit{v} for each of the query vectors,
where the weight is given by the compatibility between the query and key vectors: \textit{q} and \textit{k}. The final compatibility score is obtained after applying softmax on the pairwise dot products as described in Equation \ref{eq:qkv_att}.
\begin{equation}
    Attention(Q, K, V) = softmax(\frac{QK^T}{\sqrt{d_k}})V
  \label{eq:qkv_att}
\end{equation}
This attention mechanism is applied multiple times in parallel through a set of attention heads, each one with a distinct set of parameters, such that the attention mechanism is allowed to focus on different input subspaces. The output is finally concatenated and projected to the initial token dimension. The multihead functionality is laid out in Equations \ref{eq:mha1} and \ref{eq:mha2}.
\\
\begin{equation}\label{eq:mha1}
MultiHead(Z) =
      Concat(head_1, ..., head_h)W^O
\end{equation}
\vspace{-0.5cm}
\begin{equation}\label{eq:mha2}
    head_i = Attention(QW_i^Z, KW_i^Z, VW_i^Z)
\end{equation}

In the following, we state the equations for the visual and language features' tokenization and projection, embedding addition, and concatenation operations prior to feeding the result to the TransFusion module for a single scale level.

\begin{align}
    \quad\quad Vf &= patchify(Vf_i) \in \mathbb{R}^{N \times P^2\cdot c} \\
    Vf &= Vf W_p; W_p \in \mathbb{R}^{P^2\cdot c \times D} \\   
  Lf &= LM(X); lf \in \mathbb{R}^{L_A \times D} \\
    Vf & \mathrel{{+}{=}} Vf_{emb} + Pos_{emb} \\
    Lf &= Lf + Lf_{emb} \\
    Lf &= Dropout(Lf) \\
    Vf &= Dropout(Vf) \\
    Z &= Concat(Vf, Lf) 
\end{align}
where $Vf_{emb} \in \mathbb{R}^{D}, Pos_{emb} \in \mathbb{R}^{N \times D}$, $Lf$ represent the tokenized language features and $Vf$ represent the tokenized visual features.  

As described in Section 3, the TransFusion model consists of multiple transformer encoder layers applied in succession. This mechanism is replicated on multiple input scales to enhance the corresponding visual features. A single transformer encoder layer employs Layer Normalization \cite{layer_norm}, MLP blocks, multihead QKV self-attention, a Dropout module\cite{dropout}, and the GELU\cite{gelu} non-linearity. 
The functioning is described in equation \ref{eq:transformer_enc}, where we drop the scale level indices for simplicity.

\begin{equation}\label{eq:transformer_enc}
\begin{aligned}
    Z' &= LN(MultiHead(Dropout(Z)) + Z) \\
    Z ' &= MLP(Dropout(GELU(Z'))) + Z' \\
    Z' &= LN(Z')
\end{aligned}
\end{equation}


\section{Implementation details}
\label{sup:implementation}

For the majority of our runs, we use a learning rate of 1e-4. For training the backbone encoders, we additionally decay the learning rate by 5 to better synchronize with the fusion module dynamics that start from random initialization.

We augment the data by altering both height and width resolutions while ensuring their downsampled shapes are divisible by the patch sizes. The following height-width pairs are used for most of the experiments: 480-596, 544-640, 640-768, 704-896, 768-896, 800-1200. Before rescaling the images, they are cropped randomly in a relative range of 0.9 for both height and width. This way, we preserve about 80\% of the original visual area and reduce the chances of evicting GT boxes. The images are flipped horizontally with a probability of 50\%. We apply a moderate amount of color jittering: we alter the brightness in a relative range of $[-0.15, 0.15]$, the contrast in $[-0.1, 0.1]$, and the hue in $[-0.05, 0.05]$. For reference, Ego4D resizes the image height to 800 pixels while limiting the width to 1333 pixels.
By the choice of height and width ratios, both approaches also provide a weak form of aspect ratio augmentation.

We use a 1D sinusoidal positional embedding for the visual tokens. The patch dimensions used per level are the following: high-resolution ResNet-50 runs use patch projection sizes of $4, 4, 2, 1$ for the FPN stages. Smaller patches tend to give better performance, but increase the computational cost of the self-attention mechanism that scales with the square of the number of tokens.
We also apply $0.1-0.2$ (depending on the language model size) language token dropout and $0.1$ visual token dropout before the transformer fusion layers, which slightly improves validation performance.
Each of our models is trained on a single NVIDIA A100 GPU with 80GB of VRAM.

 During development, we observed that the \methodname architecture reaches high confidence in predicting foreground objects  with a corresponding local minimum of the loss  before learning to effectively fuse the visual and language modalities. This diminished the final classification performance for better box localization. Hence, we reduced the region proposal network's sampling batch size and the detection network's image batch size from 256 and 512 to 64 and 128 respectively. penalizing the model less for foreground-background mismatches. This reduces the dependence on visual features, which are already adapted for object detection tasks. 
We also perform multiscale augmentation by resizing the shortest edge between 480 and 800 pixels, random-relative cropping, color jittering, and image horizontal flipping to enable a longer learning stage and more effective feature fusion. 

\section{Evaluation of action contexts}
\label{sup:summaries}

\begin{table*}[!h]
    \centering
    \small
    \begin{tabular}{c|c|c|c|c|c}
    \hline
    Lang.\ used during training & Lang.\ used during inference & NO $\uparrow$ & VO $\uparrow$ & N $\uparrow$ & N-V $\uparrow$\\
    \hline\rule{0pt}{2.2ex}

    $\mathcal{A}+\mathcal{N}_s$ & next active obj.\ (GT) & \textbf{37.40} & 11.65 & \textbf{21.86} & \textbf{8.05}\\
    
    $\mathcal{A}+\mathcal{N}_s$ & $\mathcal{A}+\mathcal{N}_s$ & 33.20 & \textbf{12.00} & 19.63 & 7.36\\
    
    $\mathcal{A}+\mathcal{N}_s$ & $\emptyset$ & 29.51 & 10.72 & 17.49 & 6.46\\

    \hline \rule{0pt}{2.2ex}
    
    $\mathcal{N}_s$ & next active obj.\ (GT) & 35.42 & 9.24 & 20.78 & 7.33\\
    
    $\mathcal{N}_s$ & $\mathcal{N}_s$ & 33.33 & 10.61 & 19.62 & 7.13\\
     
    $\mathcal{N}_s$ & $\emptyset$ & 29.73 & 8.65 & 17.83 & 6.18\\
    \hline \rule{0pt}{2.2ex}

    $\emptyset$ & $\emptyset$ & 31.26 & 10.64 & 17.71 & 6.14 \\
    \hline
    
    \end{tabular}
    
    \caption{\centering \textbf{Performance of various action context combinations at train and inference time.} Comparison of performance obtained on the Ego4D validation set with models trained on $\mathcal{A}$ and $\mathcal{A} + \mathcal{N}_s$, as well as a visual-only model. For the language-aided models, we experiment with forwarding the ground-truth next active object labels, passing the intended action context to the model, and using no language input ($\emptyset$). Experimental results show that having the ground-truth NAO labels can further improve the model performance, and that even in the absence of language input ($\mathcal{A} + \mathcal{N}_s$ \& $\emptyset$), our language-aided model performs competitively to a model trained without language input ($\emptyset$ \& $\emptyset$).}
    \label{tab:nao_gt_input}
\end{table*}

\begin{table*}[!t]
    \centering
    \small
    \begin{tabular}{c|c|c|c|c|c||c}
    \hline

    & $\mathcal{A}$ & $\mathcal{N}_h$ & $\mathcal{N}_s$ &  $\mathcal{A}+\mathcal{N}_h$ & $\mathcal{A}+\mathcal{N}_s$ & GT\\
    \hline \rule{0pt}{2.2ex}
     Abs. & 4,662 & 1,119 & 5,390 & 5,184 & 7,484 & 11,576 \\
     Rel. & 27.0\% & 6.49\% & 31.3\% & 30.1\% & 43.5\% & 67.2\%\\
     $\mathrm{NO}$ $\uparrow$ & 32.41 & 31.16 & 33.33 & 32.72 & 33.20 & 37.33 \\
     \hline \rule{0pt}{2.2ex}
     $\mathrm{NO}^{+}$ $\uparrow$ & \textbf{41.31} & \textbf{47.42} & \textbf{44.24} & \textbf{40.60} & \textbf{44.60} & \textbf{43.71}\\
     $\mathrm{NO}^{-}$ $\uparrow$ & 30.52 & 30.97 & 31.09 & 31.60 & 31.22 & 30.78\\
    \hline    
    \end{tabular}
    
    \caption{\centering \textbf{Performance analysis based on the occurrence of next active object labels in language context.} Absolute (Abs.) and relative (Rel.) frequency of the occurrence of the ground-truth next active object class in different types of action context on the Ego4D validation set, together with the performance of models trained on these action context types. Performance is measured in terms of Noun-Only mAP and separately for the full validation set ($\mathrm{NO}$), the subset where the ground-truth next active object class appears in the language summary ($\mathrm{NO}^{+}$), and the subset where it does not ($\mathrm{NO}^{-}$). The results show the benefit of the next active object appearing in the language input.}
    \label{tab:gt_noun_occurrences}
\end{table*}

 \begin{figure*}[!ht]
    \begin{subfigure}{0.49\linewidth}
        \centering
        \includegraphics[width=.95\textwidth]{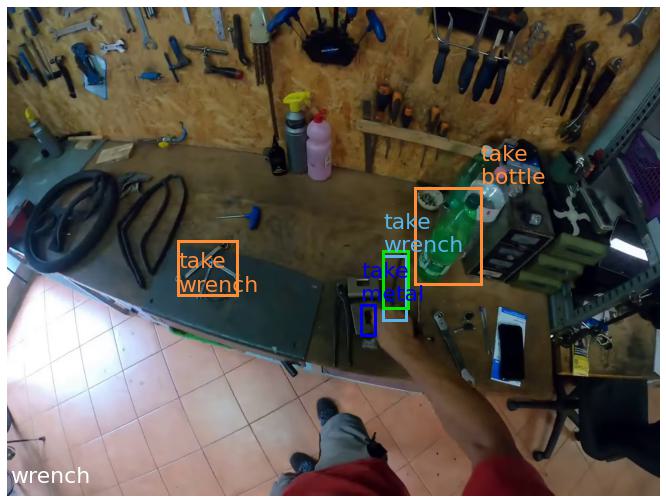}
    \end{subfigure}
    \begin{subfigure}{0.49\linewidth}
        \centering
        \includegraphics[width=.95\textwidth]{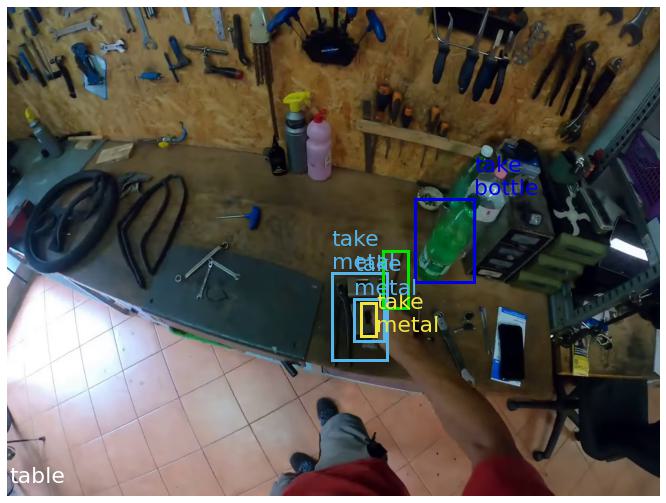}
    \end{subfigure}
    
    \begin{subfigure}{0.49\linewidth}
        \centering
        \includegraphics[width=.95\textwidth]{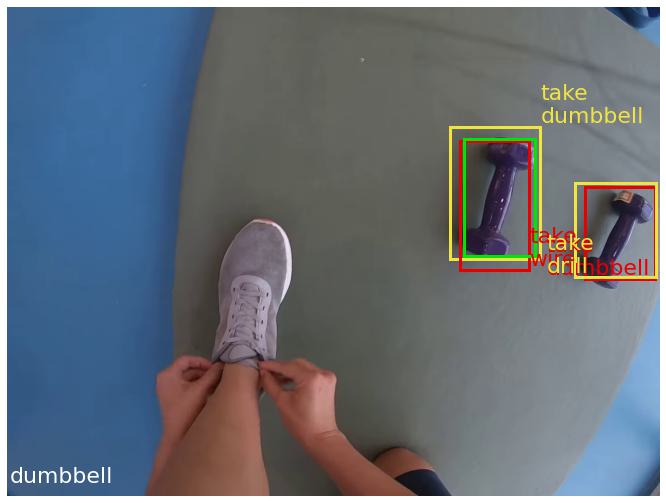}
    \end{subfigure}
    \begin{subfigure}{0.49\linewidth}
        \centering
        \includegraphics[width=.95\textwidth]{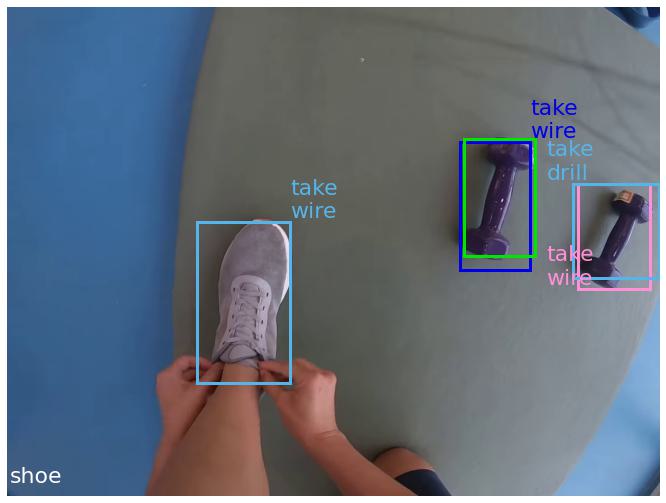}
    \end{subfigure}
    
    \caption{\textbf{Examples from the counterfactual analysis experiment.}
    \centering We show the changes in our model's predictions when altering the language input from \textit{wrench} to \textit{table} on the top row, and \textit{dumbbell} to \textit{shoe} on the bottom row. Additional, similar visualizations are available in \autoref{fig:language_change_extra}.}
    \label{fig:language_altering_1}

\end{figure*} 
    
\paragraph{Correlation of model performance with language input}

Table 2 provides a high-level comparison of the performance obtained using different types of action context language input. In \autoref{tab:nao_gt_input}, we further experiment with providing our models trained on $\mathcal{A} + \mathcal{N}_s$ and $\mathcal{N}_s$ different types of action context during inference. Specifically, we compare between (1) using the original action context class(es) each model was trained with as language input, (2) using the object to be interacted with next, taken from the ground-truth labels, as language input, and (3) an ablation where we omit all language input.

Ideally, our models trained with $\mathcal{N}_s$ and $\mathcal{A} + \mathcal{N}_s$ input should be able to make use of salient objects enumerated in the language input to better disambiguate between multiple possible next active objects (NAOs). The ground-truth NAO forms a reasonable ``best-case" version of the $\mathcal{N}_s$ inputs: we would expect an increase in the models' performance if the NAO is highlighted to the model as the only salient object in the prediction frame. Indeed, as evidenced in \autoref{tab:nao_gt_input}, both the model trained on $\mathcal{N}_s$ and that trained on $\mathcal{A} + \mathcal{N}_s$ perform better when receiving the ground-truth NAO as input, than when receiving the salient objects $\mathcal{N}_s$ from the context generation models. On the other hand, the performance drops when the models do not receive any language input. We would like to point out that the performance of the language-aided models does not suffer significantly when omitting language input, suggesting that the models learn to benefit from the provided action context rather than becoming dependent on it.

We further conduct a comparison of our model's performance on samples for which the ground-truth NAO appears in the generated language input against that on samples for which it does not and calculate the absolute and relative frequencies of the ground-truth NAO's appearance. We showcase the results in \autoref{tab:gt_noun_occurrences}. The aforementioned considerations strongly suggest that the models benefit from the $\mathcal{N}_s$ action context specifically due to its ability to highlight salient objects in the actor's environment.

\paragraph{Counterfactual analysis}
We showcase how changing the action context alters the predictions of the model on various prediction frames from the Ego4D dataset's validation set. Figure \ref{fig:language_altering_1} together with Figure \ref{fig:language_change_extra} illustrate the difference in the predictions of an $\mathcal{N}_s$-trained model when using language input consisting of the ground-truth class name (left column), compared to using the class name of another object in the image or a similar-looking object (right column). We visualize the top 4 highest-scoring bounding boxes, along with the bounding box capturing the ground-truth next active object (in green) and the language input to the model, visible on the bottom left of the images. The visualizations indicate qualitatively that TransFusion learns to effectively condition its predictions on action context encoded as language summaries to anticipate future object interactions.


\section{Comparison with state-of-the-art}
\label{sup:ego4d}

In this section, we provide more details on the experiment setup reported in the main paper. The TTC values are obtained using the provided Ego4D model checkpoint while keeping our original box, noun, and verb predictions.
For the language encoder, we use SBERT 384 and for the visual encoder, we use the frozen Ego4D ResNet-50 weights. We discount the classification loss for the background class prediction by 0.8 such that the model focuses more capacity on the actual object  categories. 

\paragraph{Validation-test performance variance}
We observe some noticeable variance between the validation and test set performance, both for our model and the Ego4D method. We believe that this is caused by multiple factors, such as 1) only one validation fold being used during the training of the two models. Performing $k$-fold cross-validation provides a more reliable estimate of the \emph{true} model performance, at the cost of a significantly larger computation time. 2) even when using a single validation fold, a reasonable performance estimate can be obtained. In our case, we identify a noticeable class distribution shift when moving from the training to the validation set. It is plausible that a similar distribution shift occurs between the validation and test set.

\subsection{Additional comparisons}
We provide additional insights on how our method performs compared to that of Ego4D, highlighting the effectiveness of our approach and the suitability of using language descriptions to summarize the action context. 

\paragraph{Model performance as a function of label distribution}
We evaluate the performance of the two models separately for the most frequent and for the tail class categories. This comparison confirms the effectiveness of our method: we register consistent improvements for both frequent and rare categories: over 15\% and 22\% for nouns and 132\%, 161\% for verbs respectively. This is very encouraging, seeing as improving performance in low-tail classes is a challenging aspect for many prediction tasks and the adoption of language descriptions could provide further breakthroughs. The results are presented in \autoref{tab:ego4d_top_ss_tail_noun_map}
 and \autoref{tab:ego4d_top_ss_tail_serb_map}. We report the classification-only results (without conditioning on a correct box prediction) because 
we want to highlight the substantial classification improvements owed to improved semantic understanding. The corresponding metrics are denoted as NO and VO respectively. 
 
 \begin{table}[!h]
    \centering
    \small
    \begin{tabular}{c|c|c|c}
    \hline
    Model & NO $\uparrow$ & Top-10 NO $\uparrow$ & Tail NO $\uparrow$\\
    \hline
    Ego4D & 28.70 & 27.58 & 26.45   \\ 
    \methodname & \textbf{33.80} & \textbf{31.84} & \textbf{32.37} \\
    \cdashline{1-4}
    Improvement & 18\% & 15\% & 22\% \\
    \hline
    \end{tabular}
    \caption{\textbf{Noun-only mAP} for top 10 and tail noun categories on Ego4D dataset. We observe consistent gains over the full class spectrum, in particular in tail classes. }
    \label{tab:ego4d_top_ss_tail_noun_map}
\end{table}

\begin{table}[!h]
    \centering
    \small
    \begin{tabular}{c|c|c|c}
    \hline
    Model & VO $\uparrow$ & Top-5 VO $\uparrow$ & Tail VO $\uparrow$\\
    \hline
    Ego4D & 5.22 & 10.11 & 4.38   \\ 
    \methodname & \textbf{12.00} & \textbf{23.54} & \textbf{11.46}\\
    \cdashline{1-4}
    Improvement & 129\% & 132\% & 161\% \\
    \hline
    \end{tabular}
    \caption{\textbf{Verb-only mAP} for top 5 and tail verb categories on Ego4D dataset. We observe consistently strong gains over the full class spectrum, in particular in tail classes.}
    \label{tab:ego4d_top_ss_tail_serb_map}
\end{table}

\paragraph{Model performance dependence on bounding box size.}We additionally present a performance comparison based on the size of the ground-truth bounding box. We divide the ground-truth labels into 3 categories according to their area, each one containing a third of the total validation samples. The results for the Box-Noun mAP metric are shown in \autoref{tab:ego4d_boxes_lms}. We again notice consistent improvements across the different categories, which show that our method is flexible enough to perform well on difficult cases such as small bounding boxes or tail classes. This is encouraging because improving on these corner cases usually requires a lot of effort. The TransFusion model using language action context summaries achieves this without any special design choices aimed towards improving this objective. 

With respect to the Box-Noun-Verb mAP, we notice regular gains of about 1.5 points over the Ego4D methods (hence, this result is omitted from the table). This suggests that the verb label prediction is more dependent on understanding the scene and action context in a holistic manner.

\begin{table}[!h]
    \centering
    \small
    \begin{tabular}{c|c|c|c}
    \hline
    Model & Ns $\uparrow$ & Nm $\uparrow$ & Nl $\uparrow$\\
    \hline
    Ego4D & 6.39 & 15.20 & 21.82   \\ 
    \methodname & \textbf{7.91} & \textbf{18.65} & \textbf{22.75} \\
    \cdashline{1-4}
    Improvement & 23\% & 22\% & 4\% \\
    \hline
    \end{tabular}
    \caption{\textbf{Box-Noun mAP} for small (Ns), medium (Nm) and large (Nl) boxes. Box-Noun-Verb mAP improvement is about 1.5 points for each of the categories hence it is omitted. Our method improves detection and classification results on difficult cases with small bounding boxes, complementing the hard-to-encode visual cues.}
    \label{tab:ego4d_boxes_lms}
\end{table}

\section{Architecture ablation}
\label{sup:architecture}

\subsection{Learning scale-specific features} 
\paragraph{Shared multiscale fusion weights.}
An alternative take to combat overfitting is to share model weights over multiple similar inputs, such that they are forced to learn more general data representations. We apply this principle by resharing the transformer fusion weights over the multiple input scales. Because this setting imposes additional constraints on the fusion weights, we increase the fusion module's capacity by 33\% (such that it still fits on one of our GPUs) and reduce the L2 weight decay to 3e-5 to allow more optimization freedom. Using this setup, we register a notable decrease in performance. We perform multiple runs, but do not manage to score more than $7$ MAP Box-Noun-Verb and $19.2$ MAP Box-Noun on the validation set (approximately $.5$ and $1$ absolute point difference). This indicates that using shared fusion weights at multiple scales is counterproductive for our task and that the model learns different representations at different feature map scales, all needed for effective prediction. The other hyperparameters are kept fixed as in \autoref{sup:implementation}, with a context length of 3 used for $\mathcal{A}$.

\paragraph{Forwarding language features.} Instead of copying the language encoder outputs for every scale-level fusion, we forward the corresponding fused tokens to the latter stages. Additionally, we also append residual connections between the language features tokens at the successive multiscale levels to allow the model to more easily retain some of the initial information. The results are presented in \autoref{tab:feature_forwarding} where we notice the improved results obtained with the "copy" strategy as opposed to reusing features from previous scale levels. This further supports our hypothesis that each scale level operates with distinct, specialized representations and that reusing ones from different scales are detrimental to model performance. For this experiment, we use the parameters described in \autoref{sup:implementation} with a context length of 3, including salient objects.

\subsection{Language modeling loss}
We also investigate including an additional language model loss to further accelerate the action description learning. Specifically, we include an additional loss term, $\mathcal{L}_{lm} = \frac{1}{2}(\mathcal{L}_{lm_N} +\mathcal{L}_{lm_s})$ in the final optimization objective
\begin{equation}
\mathcal{L} = \mathcal{L}_{box} + \mathcal{L}_{noun} + \mathcal{L}_{verb} + \mathcal{L}_{ttc} + \mathcal{L}_{lm} 
\end{equation} The target categories for this loss are the ground-truth noun and verb labels; $\mathcal{L}_{lm_N}$ and $\mathcal{L}_{lm_s} $ are regular cross-entropy losses. The difference between $\mathcal{L}_{noun}$, $\mathcal{L}_{verb}$ and $\mathcal{L}_{lm}$ is that the latter term is applied on the mean-pooled \textit{fused language tokens} at each multiscale fusion level (i.e. the transformer encoder outputs). The former ones work on the ROI-pooled bounding-box features in the Faster R-CNN prediction heads. The experimental results however did not show any improvement when including $\mathcal{L}_{lm}$, which instead decreased the final Box-Noun-Verb performance. We believe that this task in itself can be quite challenging. Without considerably increasing the model capacity, performing it concomitantly with the detection-based objectives can negatively impact the target model's performance.

\section{Video features and context length}
\label{sup:video_features}

\subsection{Computational cost analysis}

\begin{figure}[h!]
    \vspace{-0.3cm}
        \centering
        \includegraphics[width=.85\linewidth, trim={1.5cm, 0.2cm, 0.2cm, 0cm}]{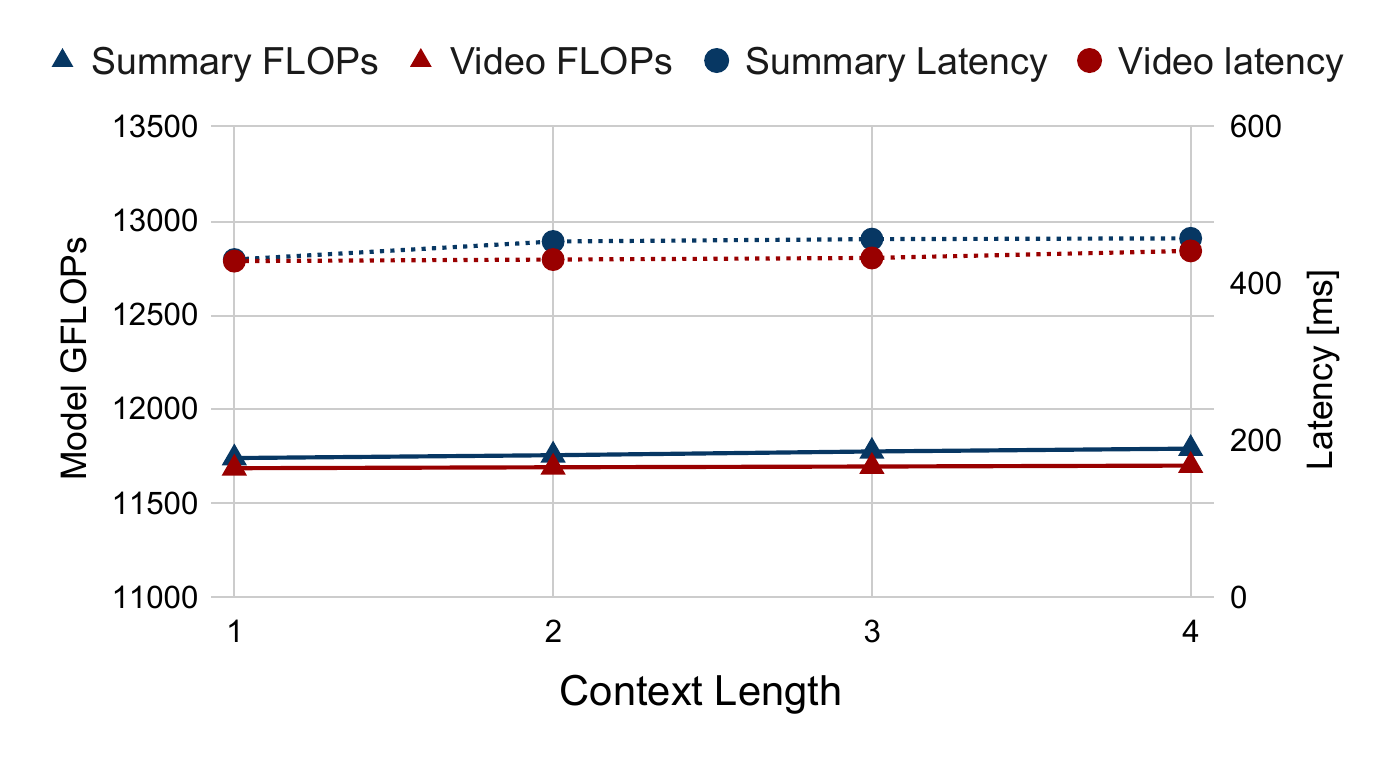}
        \vspace{-0.3cm}
    \caption{\textbf{Computational cost comparison to video features.} Comparison between TransFusion using context description summaries (Summary) and TransFusion-Video (Video). The differences between the two methods are negligible and they scale similarly with the input context length.
    }
    \label{sup_fig:comp_cost_cl}
    
\end{figure}

\autoref{sup_fig:comp_cost_cl} illustrates the inference computational cost of \methodname using action context summaries and \methodname-Video and their marginal difference. 
Given a video clip of one second, language can summarize it in two words whereas corresponding video features take up more than 12 times space (e.g. smaller SBERT feature of $2\times 384$ compared to SlowFast feature of $4\times2304$). Besides, the Ego4D 2\textsuperscript{nd} stage
SlowFast model has 33 million parameters and is trained end-to-end while the small SBERT encoder has 22 million parameters out of which we finetune only 1.7 million.  for $L_c$=3, the \methodname model has 122M trainable parameters, 11777 GFLOPs and an inference latency of 457 ms. \methodname-Video has 124M parameters, 11697 GFLOPs and an inference latency of 433 ms. Training costs reveal a similar picture, with ~2000 GFLOPs and ~250ms latency increase for both models. 
Overall, \methodname obtains a 21\% increase in N-V performance for a similar computational budget.

\subsection{Additional comparison details}
To further confirm our findings, we perform an additional run with a video context length of 6, which covers more than 100 frames before the prediction moment. We find that this does not bring any additional performance boost compared to the runs presented in Figure 5 from the main section. It registers a top-5 Box-Noun mAP performance of 18.83 (more than 1 point below the average summary-based run) and 6.40 Box-Noun-Verb mAP (also about 1 point lower).
This generally agrees with results presented by the Ego4D paper in their Table 39, where the improvements from increasing the number of SlowFast clips suffer slightly diminishing returns. While this work is not an exhaustive comparison of different types of video feature extractors, SlowFast features are still considered effective enough to be used for the latest works, including the Ego4D 2nd-stage, hence we consider them a relevant baseline. Finally, when contrasted with the language encoder run without finetuning presented in Table 3, the difference in Box-Noun-Verb mAP is 0.8 points (or about 12\%). 
This indicates the benefit of using language descriptions even when not finetuning the language encoder. 
To perform the context length comparison, we use the default parameters presented in \autoref{sup:implementation}.

\section{EPIC-KITCHENS 100}
\label{sup:epic}

\paragraph{Dataset preparation}
The SlowFast features for the EK100 dataset are extracted with an 8x8 ResNet-101 SlowFast backbone pre-trained on the Kinetics 400 dataset \cite{kinetics400}, using a stride of 16 frames and a window size of 32 frames. This replicates the extraction procedure for the Ego4D-provided features to ensure comparability. We emphasize that the EK100 dataset only provides low-resolution images of 256x456 (that we upsample to 480x640; sizes given in height-first notation) which makes the prediction task even more difficult. A certain degree of noise in the ground-truth bounding boxes generation can be expected. In some cases, there are multiple matching object detections; hence we use the model confidence score to break ties and choose the bounding box with the largest value. 

On average, the EK100 scenes are much more cluttered and more difficult to interpret than the Ego4D ones because of the several objects that could be interacted with in a kitchen. Kitchen scenes represent only 11\% of the Ego4D dataset and we do not expect our model to excel in those environments more than the others, hence we consider EK100 to be a different benchmark.

\paragraph{Evaluating the generated action context features}
\autoref{sup_tab:ego4d_vs_ek_captions} shows a comparison of the action contexts generated for EK100 and Ego4D datasets. We show the exact number of noun and verb hits (i.e. how often the generated descriptions contain the ground-truth noun-verb annotations) and the noun and verb GloVe similarity (i.e. the dot product between the generated and ground-truth noun/verb 300-d GloVe vectors). On average, according to the presented metrics, the context descriptions generated for Ego4D are closer to the respective ground-truth annotations, which translates into better overall performance. 
A large decrease in the caption quality is due to the decrease in input image resolution. The models we rely on such as OFA \cite{ofa_captioning} or UniDet \cite{unidet} are known to perform better with larger input resolutions. We quantify this decrease in performance by downsampling Ego4D from 1920x1080 to 456x256 to match the EK100 resolution and comparing the generated $N_s$ versions. We register a 23\% decrease in exact hits and a 10\% decrease in average GloVe similarity, which are significant enough to reduce the performance of our system.

\begin{figure}
    \centering
    \includegraphics[width=\linewidth]{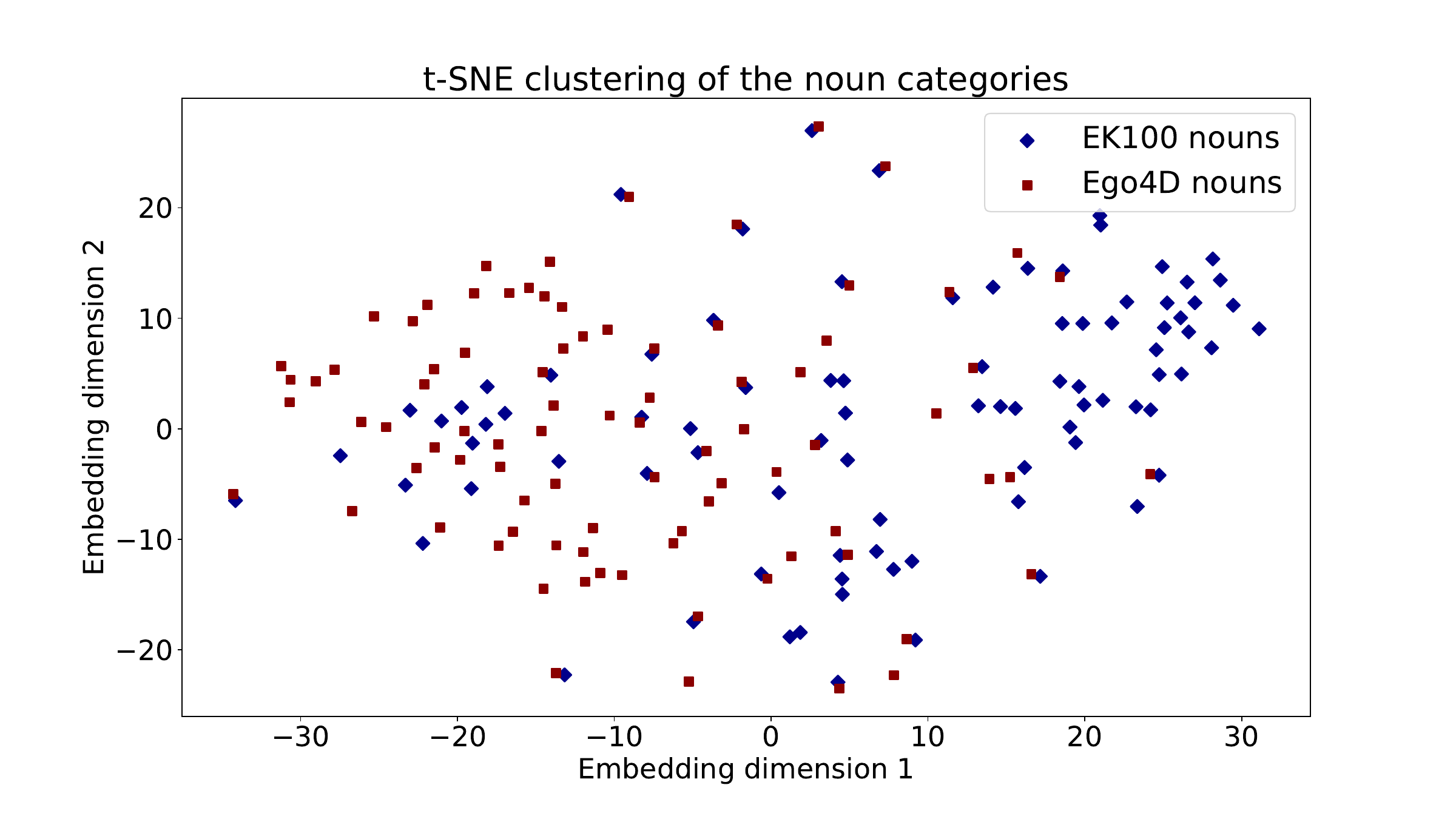}
    \caption{\textbf{t-SNE embedding of the noun categories for the two datasets.} A sizeable cluster of the EK100 nouns can be noticed on the right, while the Ego4D noun categories are more grouped towards the left. This shows the domain gap between the two datasets for noun categories.}
    \label{fig:noun_tsne}
\end{figure}

\begin{figure}
    \centering
    \includegraphics[width=\linewidth]{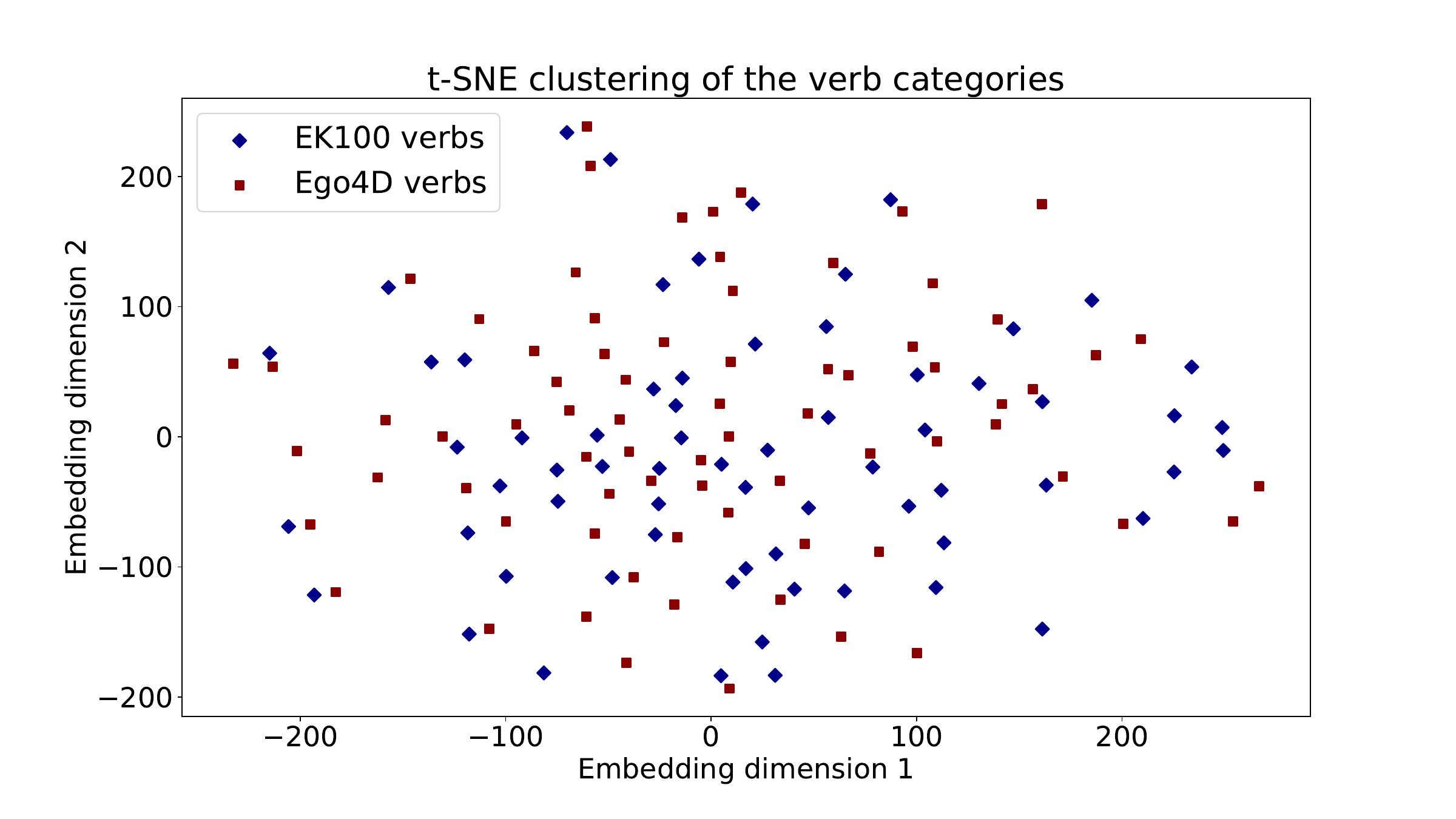}
    \caption{\textbf{t-SNE embedding of the verb categories for the two datasets}. The verbs are more uniformly spread and cover approximately the same space, which makes transfer learning easier.}
    \label{fig:verb_tsne}
\end{figure}

\paragraph{Language encoder choice}
According to our experiments, the choice of language encoder is also important for EK100, where the larger SBERT/RoBERTa language encoder seems to be better suited. We hold that this is due to the fine class differences that must be understood by the model: salad vs.\ parsley, drink vs.\ bottle, bean vs.\ nut, and lemon vs.\ fruit. This requires a more expressive and powerful language encoder to pick up the finer details, whereas a coarse understanding was enough for Ego4D, since its classes are more semantically distinct. The results shown in \autoref{sup_tab:ek_vs_ego4d_lang_encoder} confirm this hypothesis. We obtain a 0.15 Noun-Verb increase when switching from the small SBERT/BERT to SBERT/RoBERTa version and a .6 Noun increase. Compared to Ego4D, this represents an overall improvement of over 10\% in Noun mAP when using the larger language encoder. Especially for the noun domain these differences are representative and suggest that correctly predicting the noun category is more difficult on EK100 than on Ego4D. 

\begin{table}[t]
    \centering
    \small
    \begin{tabular}{c|c|c|c}
    \hline
    Dataset & Language Encoder & N $\uparrow$& N-V $\uparrow$\\
    \hline
    \multirow{2}{*}{Ego4D} & SBERT/BERT &  \textbf{20.19} & 7.55 \\ 
     & SBERT/RoBERTa  &  19.78  & \textbf{7.78} \\ 
    \hdashline
     & Rel difference & -2\% & 3\% \\
     \hline
   \multirow{2}{*}{EK100} & SBERT/BERT &  6.52  & 3.72 \\
     & SBERT/RoBERTa & \textbf{7.03} & \textbf{3.8} \\
    \hdashline
     & Rel difference & 8\% & -2\% \\   
    \hline
    \end{tabular}
    \caption{\textbf{Language encoder comparison.} We compare the effectiveness of SBERT using BERT and RoBERTa backbones on Ego4D and EK100 datasets. Overall using a larger language encoder is more beneficial for EK100 because of the fine grained domain.
    }
    \label{sup_tab:ek_vs_ego4d_lang_encoder}
    \vspace{-0.1cm}
\end{table}


To support our claims, we visualize the clustering of the noun and verb categories after applying the t-SNE dimensionality reduction method \cite{tsne} on the corresponding 300-dimensional GloVe \cite{glove} embeddings. In \autoref{fig:noun_tsne}, we notice a sizeable cluster of the EK100 nouns on the right-hand side, while the Ego4D nouns tend to be clustered on the left-hand side. 
We also notice the small distance in the embedding space for the noun classes (especially the EK100), which reinforces the difficulty of separating them. Verb clustering is presented in \autoref{fig:verb_tsne}, where we can observe a much larger overlap between the two datasets' classes and theoretically, easier separability given the larger distances between them.

\begin{table*}[!h]
    \centering
    \begin{tabular}{c|c|c|c|c|} 
        &  Exact Noun Hits &  Exact Verb Hits & Noun GloVe similarity & Verb GloVe similarity \\
        \hline
       Ego4D & 22\% & 8.6\% & 0.483 & 0.575 \\
       EK100 & 12.3\% & 12.7\% & 0.386 & 0.524 \\
    \end{tabular}
    \caption{\textbf{Action context language features comparison between EK100 and Ego4D}. On average, according to the presented metrics, the context descriptions generated for Ego4D are closer to the respective ground-truth annotations, which translates into better overall performance. }
    \label{sup_tab:ego4d_vs_ek_captions}
\end{table*} 
\begin{figure*}[!h]
    \centering
    \begin{tabular}{@{\hspace{0.75\tabcolsep}} c @{\hspace{0.75\tabcolsep}} c @{\hspace{0.75\tabcolsep}} c @{\hspace{0.75\tabcolsep}} c}
      & open drawer, 0.1s & paint wall, 2.33s & take phone, 1.0s \\
    \raisebox{1.8cm}{Ego4D}
    &
    \includegraphics[width=.30\linewidth]{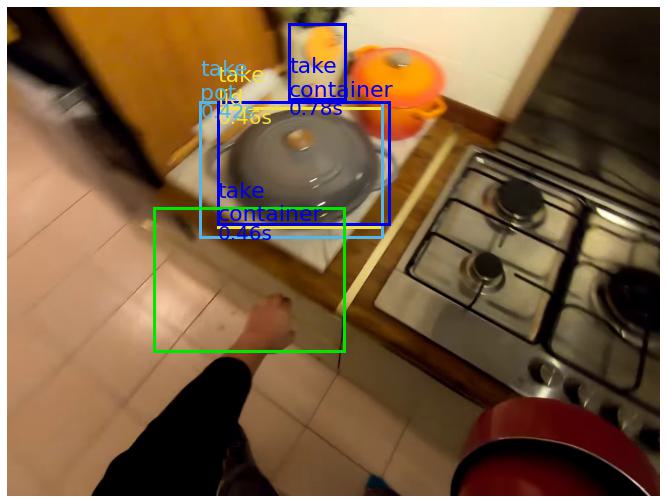}
     &  \includegraphics[width=.30\linewidth]{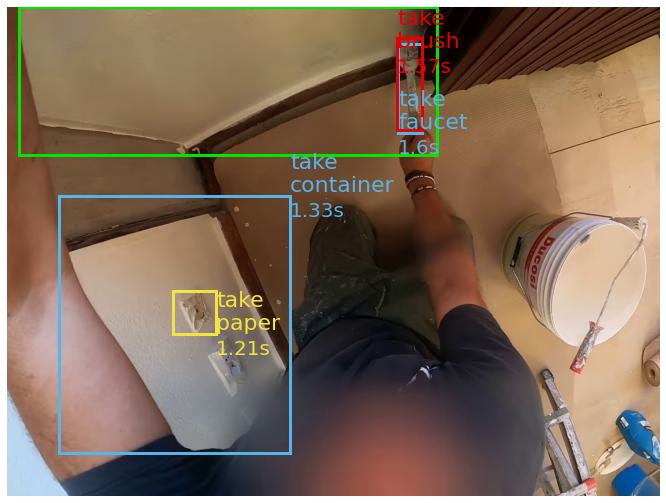} 
     &  \includegraphics[width=.30\linewidth]{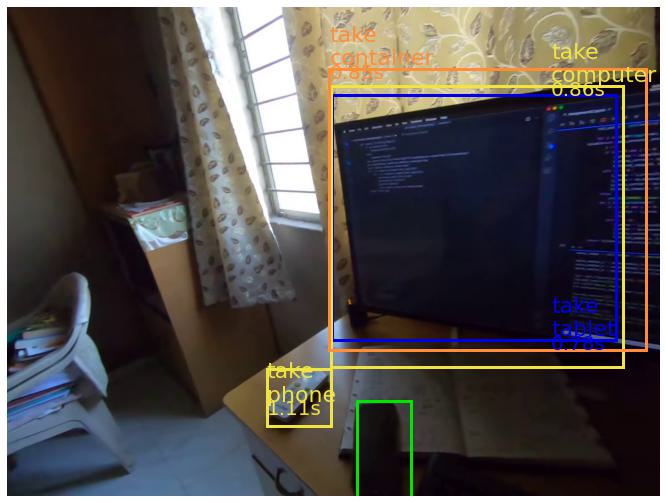} \\

     \raisebox{1.8cm}{Ours}
     &
      \includegraphics[width=.30\linewidth]{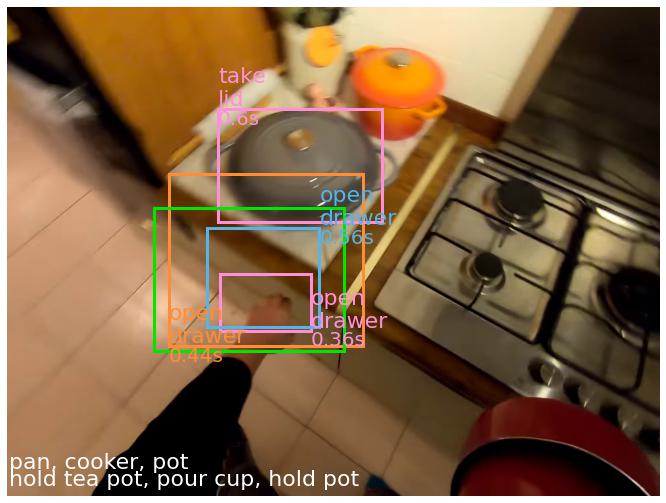}
    &  \includegraphics[width=.30\linewidth]{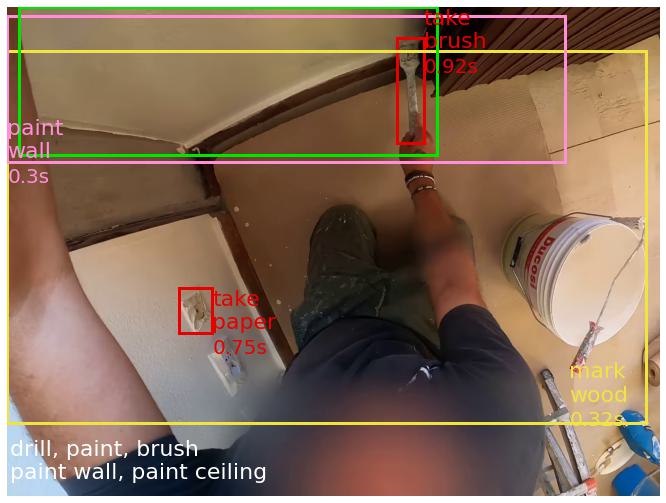}
    &   \includegraphics[width=.30\linewidth]{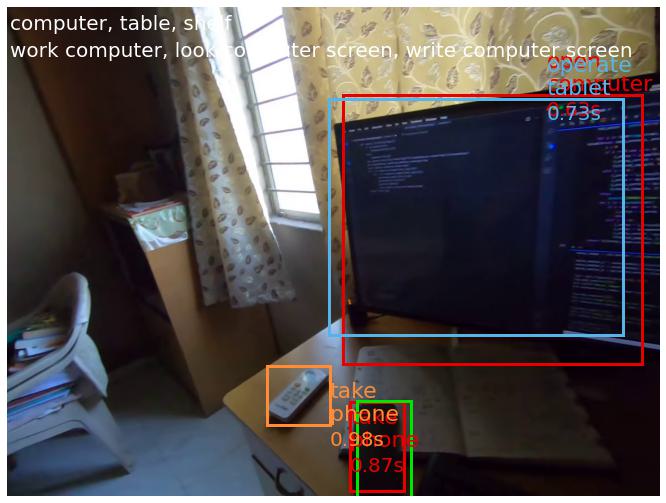}
    \end{tabular}
    \caption{\textbf{Qualitative examples of Ego4D and TransFusion (ours) predictions}. The ground-truth action label and TTC is represented on top of each column. The bright green bounding boxes denote the ground-truth location of the next object interaction. Action contexts used as language input in TransFusion are shown at the bottom in white. On average, our model manages to get more accurate predictions.}
    \label{sup_fig:qual_1}
\end{figure*}

\section{Qualitative results}
\label{sup:qualitative}
\begin{figure*}[!ht]
    \centering
    \begin{tabular}{@{\hspace{0.75\tabcolsep}} c @{\hspace{0.75\tabcolsep}} c @{\hspace{0.75\tabcolsep}} c @{\hspace{0.75\tabcolsep}} c}
    & take wire, 1.07s &  take metal, 1.17s & sand wood, 1.23s \\ 
    \raisebox{1.85cm}{Ego4D} &
     \includegraphics[width=.3\linewidth]{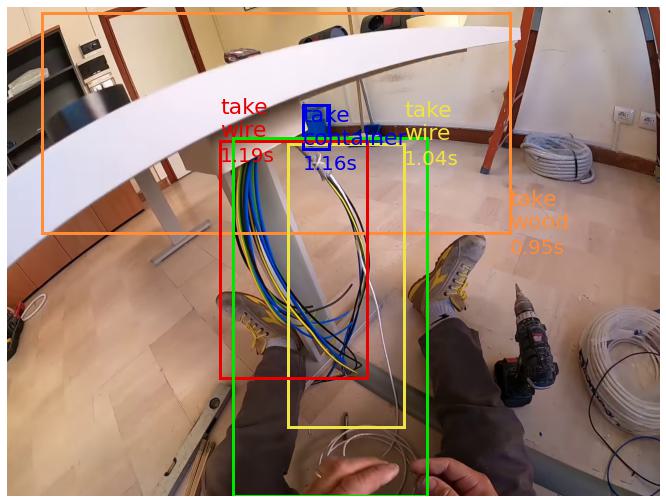}  & \includegraphics[width=.3\linewidth]{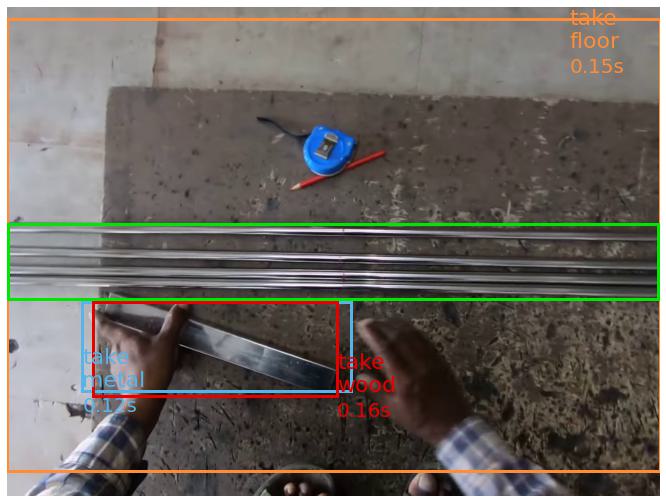} 
     & \includegraphics[width=.3\linewidth]{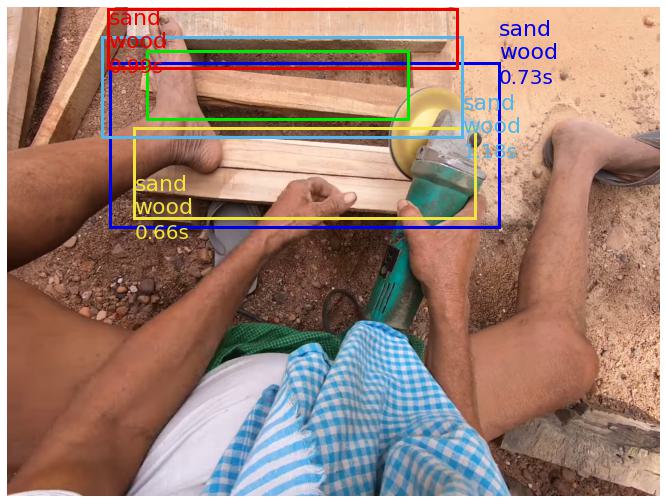} \\
         
    \raisebox{1.85cm}{Ours} &
    \includegraphics[width=.3\linewidth]{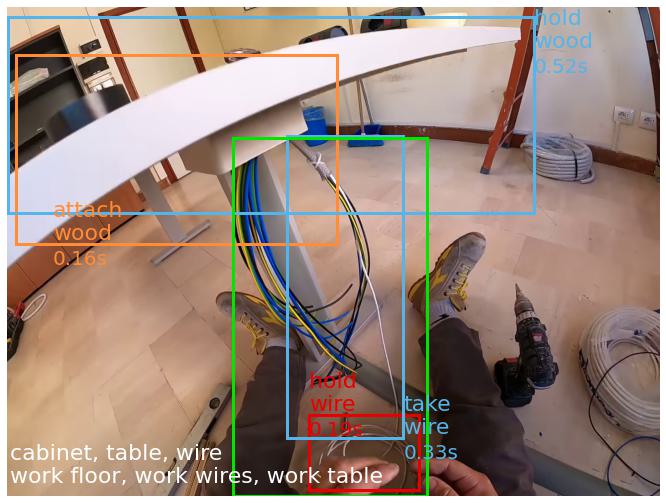} 
    & \includegraphics[width=.3\linewidth]{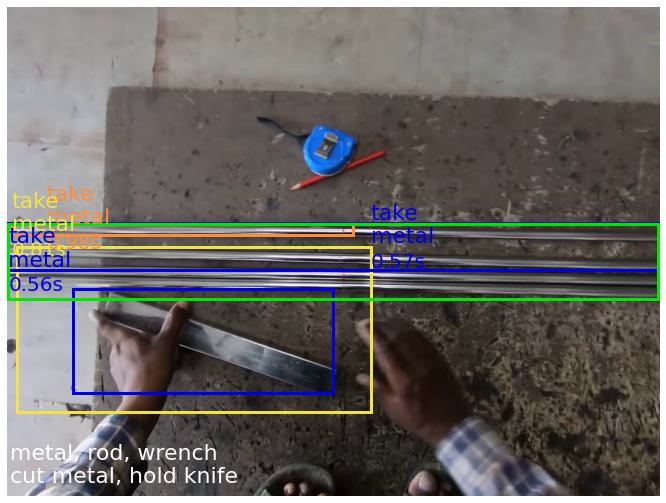}
    & \includegraphics[width=.3\linewidth]{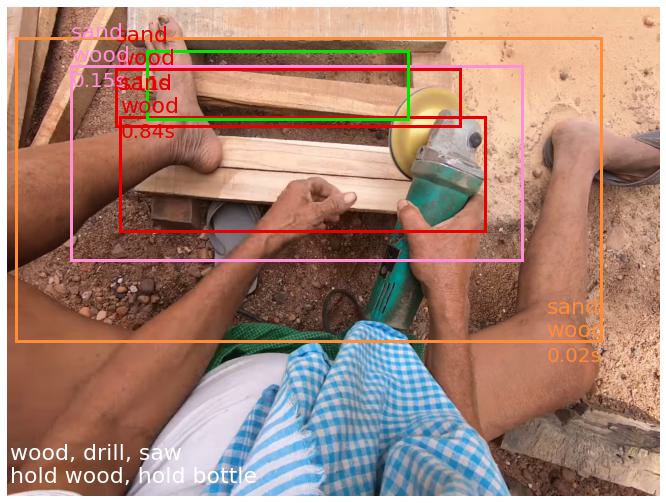} \\
        
       &  take screwdriver, 0.63s & dip brush, 0.43s & take wood, 1.53s \\ 
    \raisebox{1.85cm}{Ego4D} 
     & \includegraphics[width=.3\linewidth]{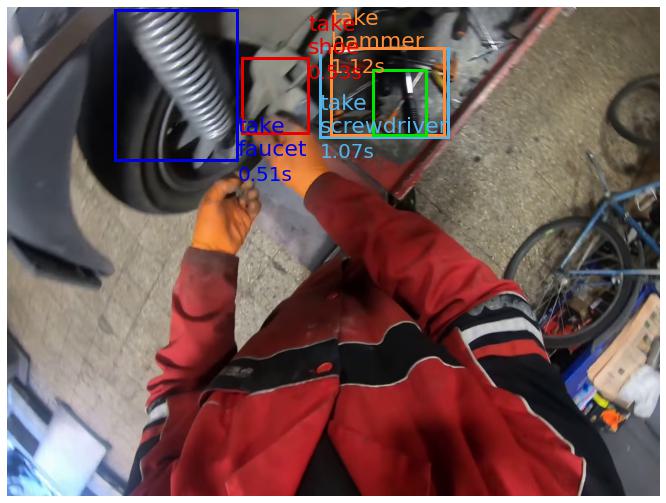} 
     & \includegraphics[width=.3\linewidth]{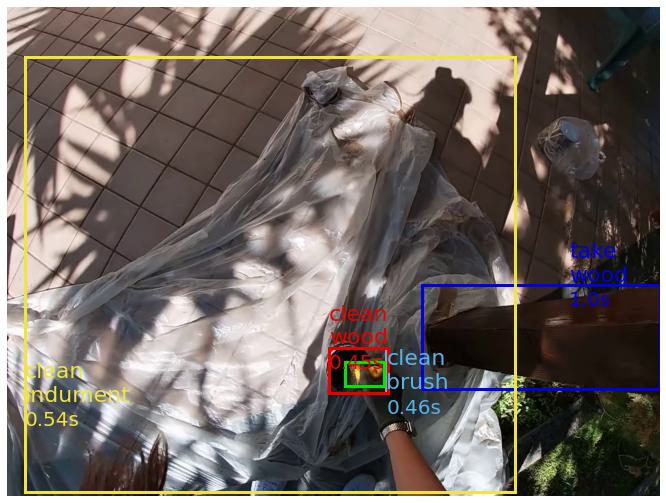} 
     & \includegraphics[width=.3\linewidth]{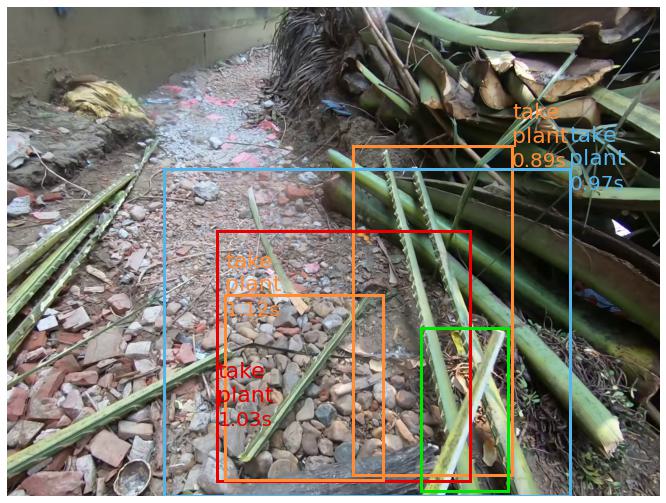} \\
     
    \raisebox{1.85cm}{Ours}
    & \includegraphics[width=.3\linewidth]{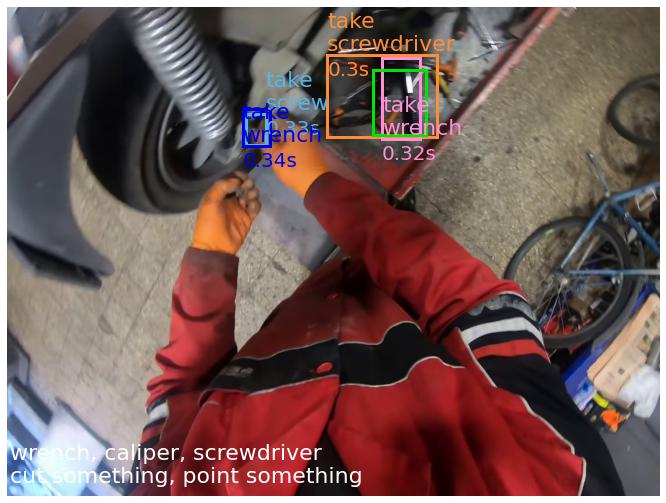}
    & \includegraphics[width=.3\linewidth]{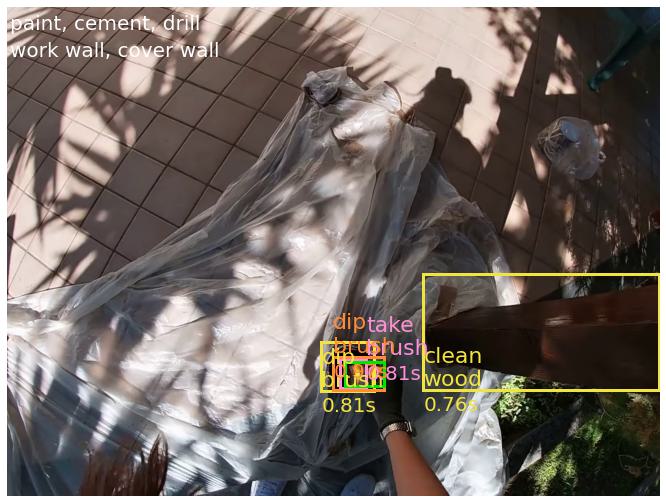}
    & \includegraphics[width=.3\linewidth]{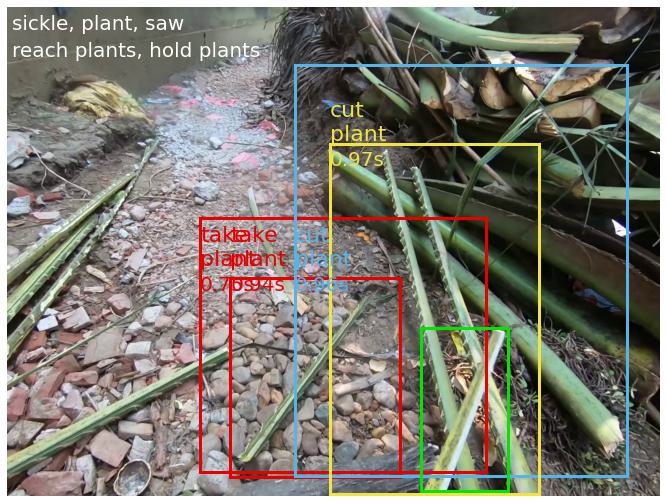} \\
   
    \\
    \end{tabular}
    \caption{\textbf{Additional qualitative examples of Ego4D and TransFusion (ours) predictions (I)}. The ground-truth action description and TTC are represented at top of each column. The bright green bounding boxes denote the ground-truth location of the next object interaction. Context summaries used in TransFusion are shown at the bottom in white. On average, our model manages to get more accurate predictions. More examples are shown in \autoref{sup_fig:qual_3}.}
   \label{sup_fig:qual_2}

\end{figure*}

\begin{figure*}[!ht]
    \centering
    \begin{tabular}{@{\hspace{0.75\tabcolsep}} c @{\hspace{0.75\tabcolsep}} c @{\hspace{0.75\tabcolsep}} c @{\hspace{0.75\tabcolsep}} c}
      & put screw, 1.63s & mask iron, 0.3s & tie wire, 0.2s \\ 
     \raisebox{1.85cm}{Ego4D} &
     \includegraphics[width=.29\linewidth]{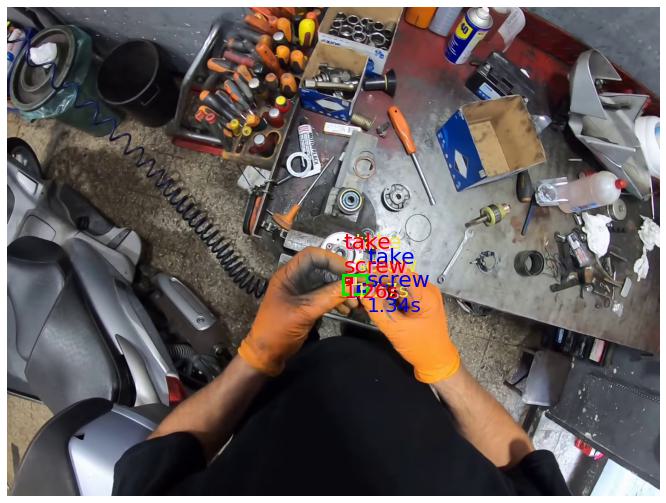} 
     & \includegraphics[width=.29\linewidth]{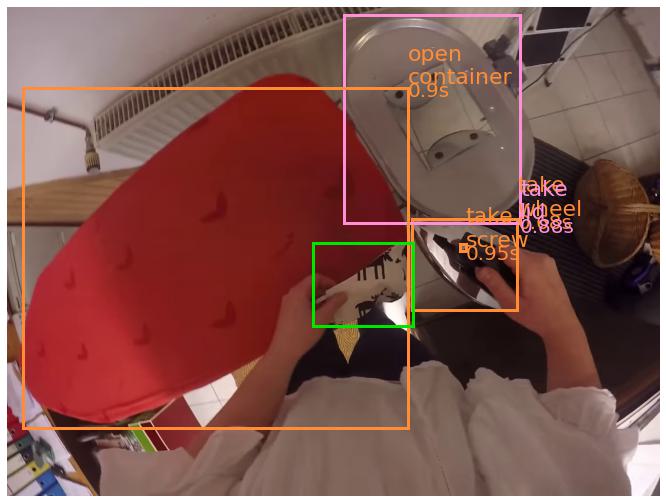} 
     & \includegraphics[width=.29\linewidth]{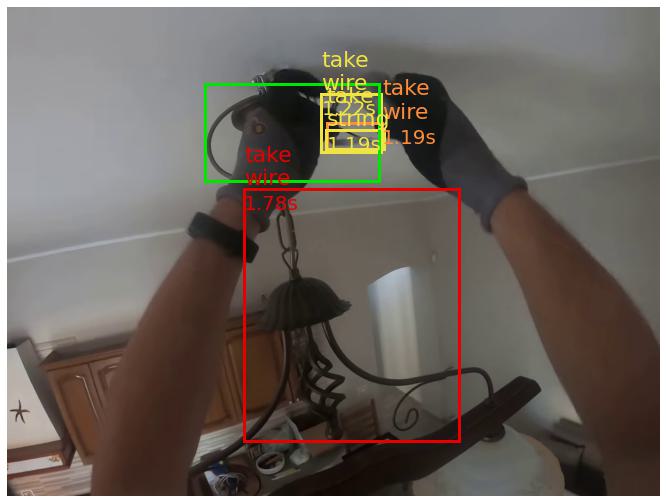} \\

    \raisebox{1.85cm}{Ours} &
    \includegraphics[width=.29\linewidth]{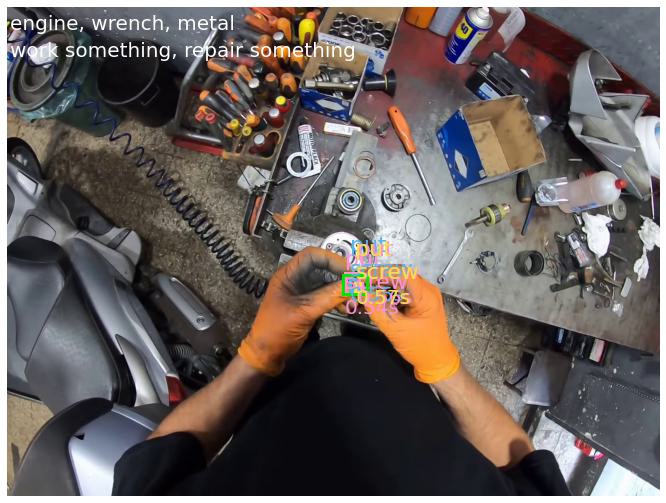} 
    & \includegraphics[width=.29\linewidth]{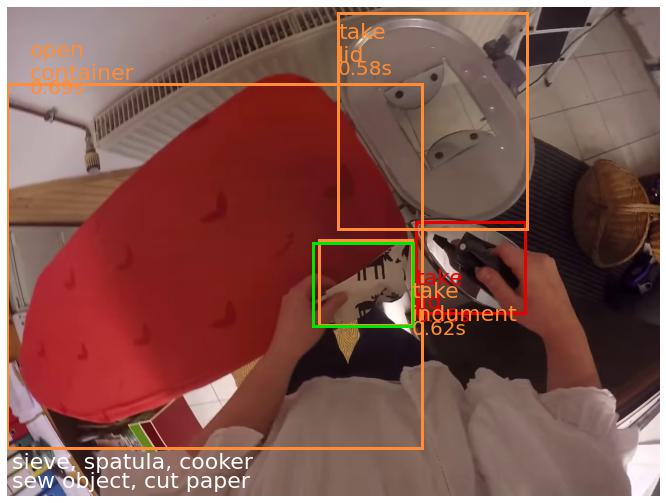}
    & \includegraphics[width=.29\linewidth]{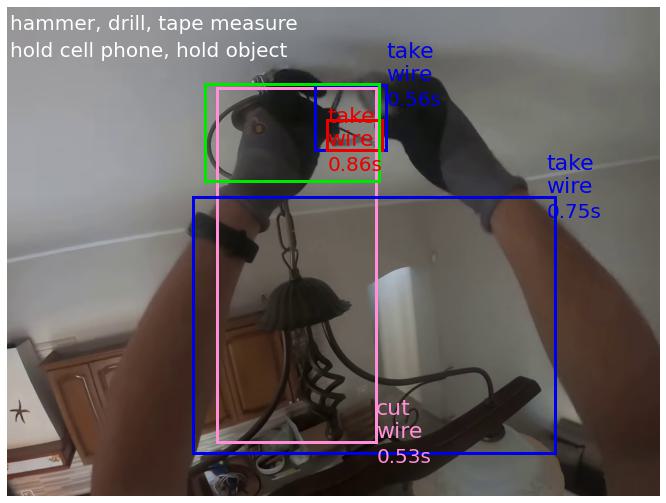} \\

    & cut plant, 1.67s & open cabinet, 0.3s & smooth wall, 1.1s \\ 
      
    \raisebox{1.85cm}{Ego4D} 
     & \includegraphics[width=.29\linewidth]{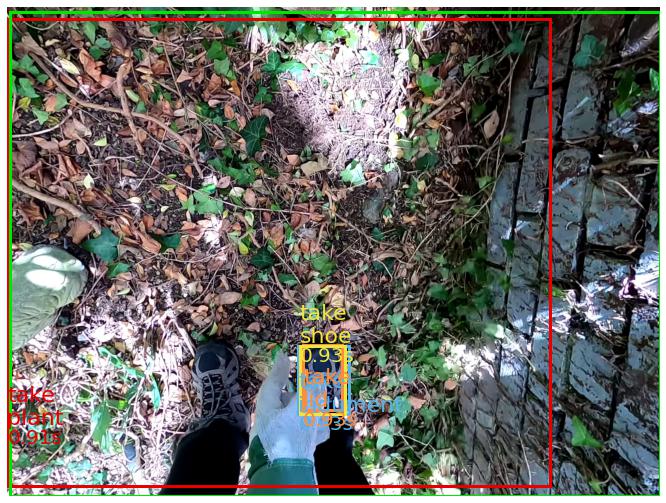}
     & \includegraphics[width=.29\linewidth]{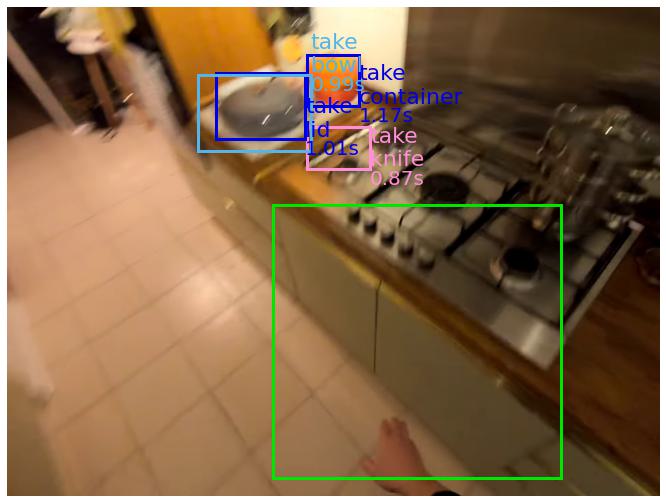} 
     & \includegraphics[width=.29\linewidth]{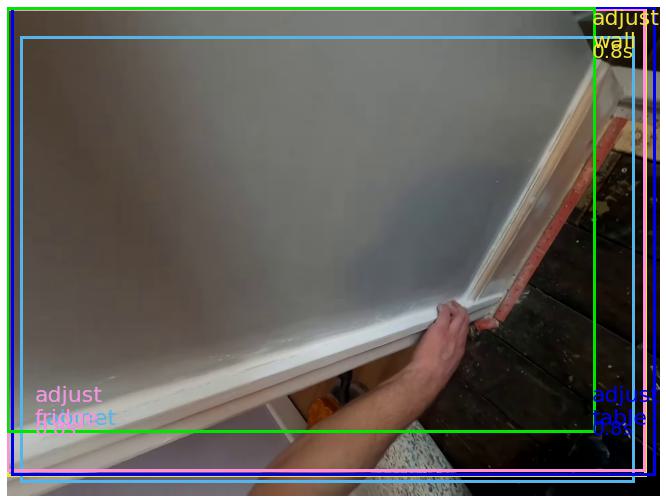} \\

    \raisebox{1.85cm}{Ours}
    & \includegraphics[width=.29\linewidth]{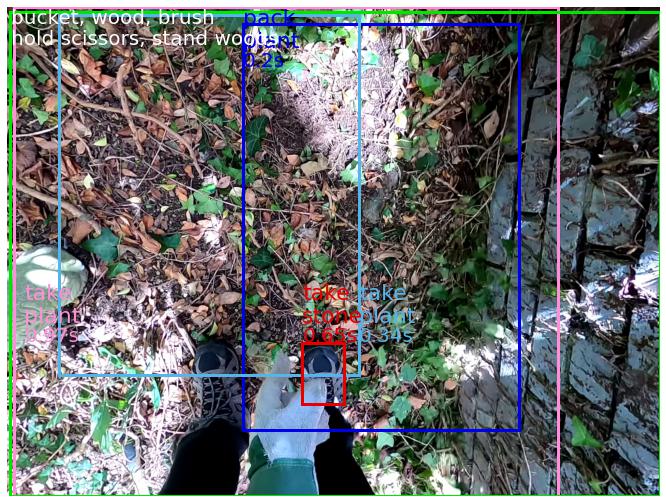}
    & \includegraphics[width=.29\linewidth]{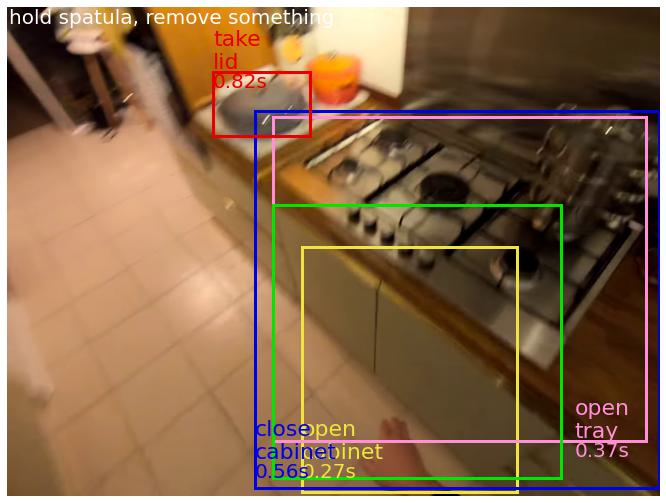} & \includegraphics[width=.29\linewidth]{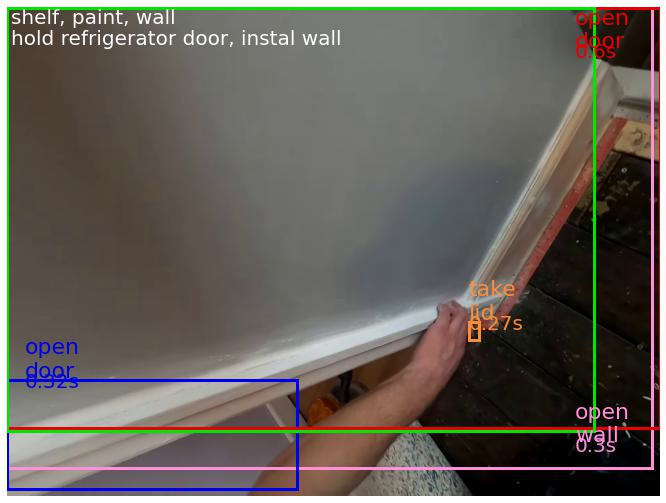} \\
    \\

    \end{tabular}

    \caption{\textbf{Additional qualitative examples of Ego4D and TransFusion (ours) predictions (II)}. See \autoref{sup_fig:qual_2} for details.}
   \label{sup_fig:qual_3}
\end{figure*}

\begin{figure*}[!ht]
    \centering
    \begin{tabular}{@{\hspace{0.75\tabcolsep}} c @{\hspace{0.75\tabcolsep}} c @{\hspace{0.75\tabcolsep}} c}
    \multicolumn{1}{p{5cm}}{\centering Changing the model predictions by changing the language input from \textit{brush} (top) to \textit{paper} (bottom)} 
    & 
    \multicolumn{1}{p{5cm}}{\centering Changing the model predictions by changing the language input from \textit{vegetable} (top) to \textit{knife} (bottom)} 
    & 
    \multicolumn{1}{p{5cm}}{\centering Changing the model predictions by changing the language input from \textit{car} (top) to \textit{paint} (bottom)} \\
    
     \includegraphics[width=.3\textwidth]{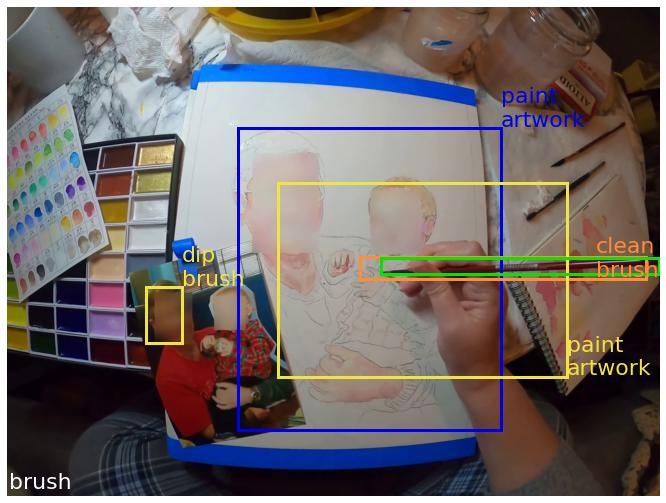}
     & 
     \includegraphics[width=.3\textwidth]{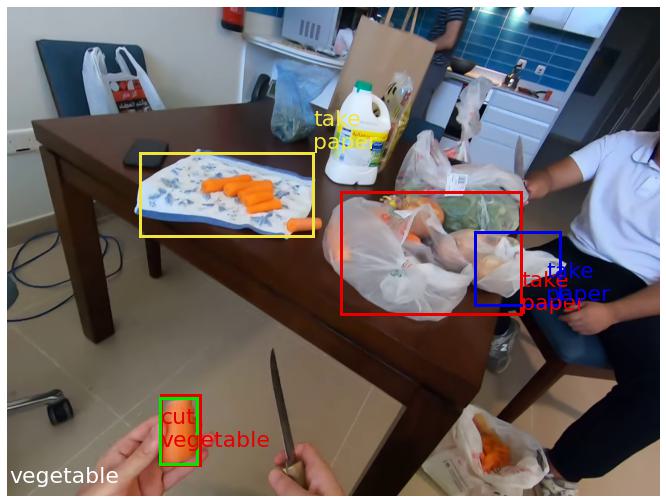}
     & 
     \includegraphics[width=.3\textwidth]{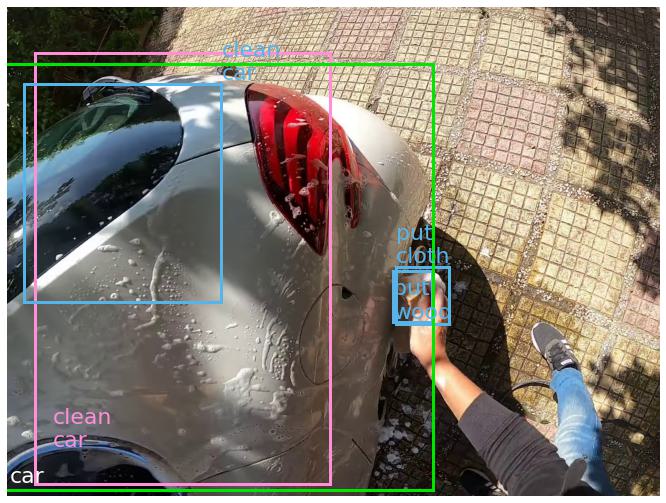} \\
     
     \includegraphics[width=.3\textwidth]{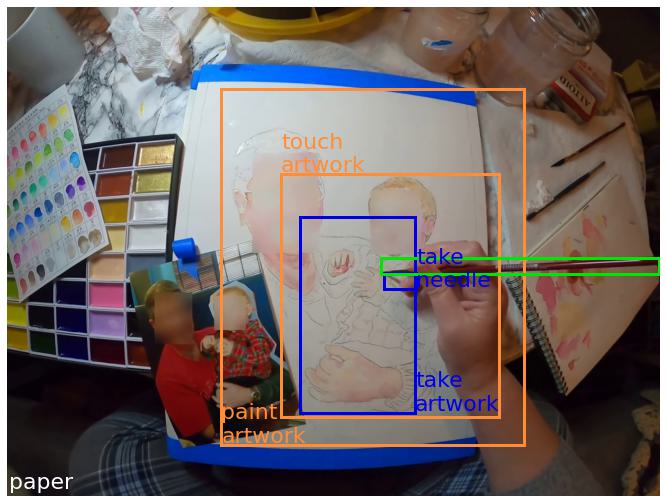}
     &
     \includegraphics[width=.3\textwidth]{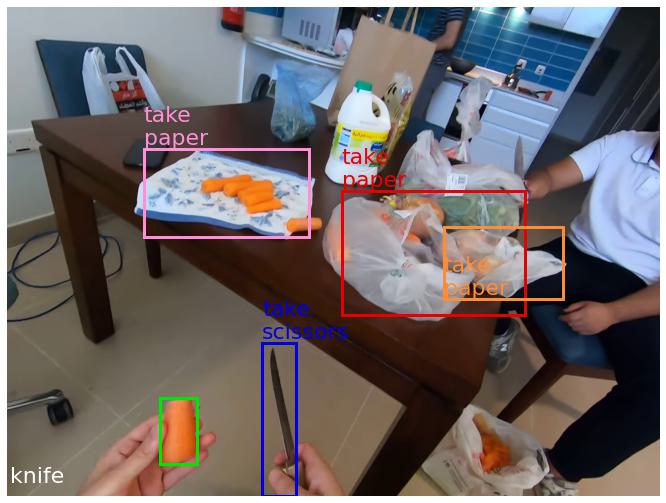}
     & 
     \includegraphics[width=.3\textwidth]{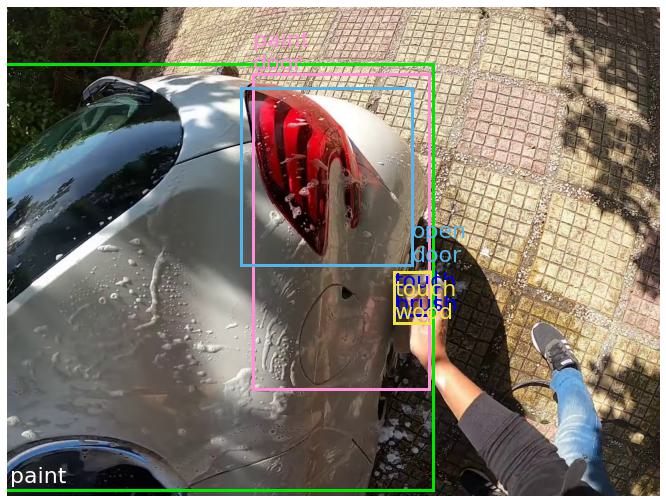} \\
    
    
    \multicolumn{1}{p{5cm}}{\centering Changing the model predictions by changing the language input from \textit{tablet} (top) to \textit{computer} (bottom)} 
    & 
    \multicolumn{1}{p{5cm}}{\centering Changing the model predictions by changing the language input from \textit{bicycle} (top) to \textit{wrench} (bottom)} 
    & 
    \multicolumn{1}{p{5cm}}{\centering Changing the model predictions by changing the language input from \textit{dough} (top) to \textit{food} (bottom)} \\
     \includegraphics[width=.3\textwidth]{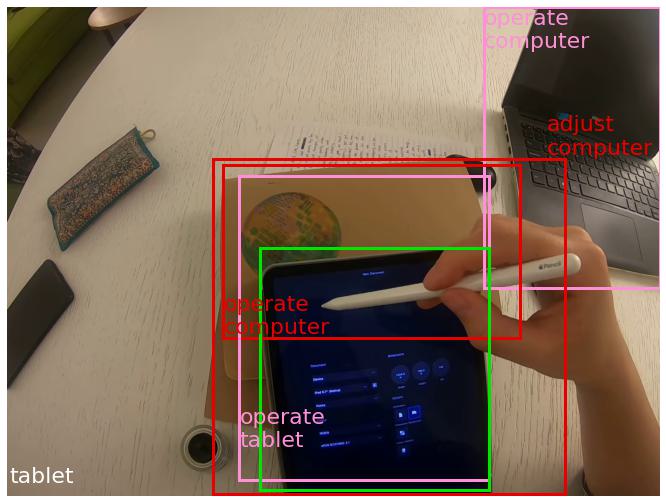}
     &
      \includegraphics[width=.3\textwidth]{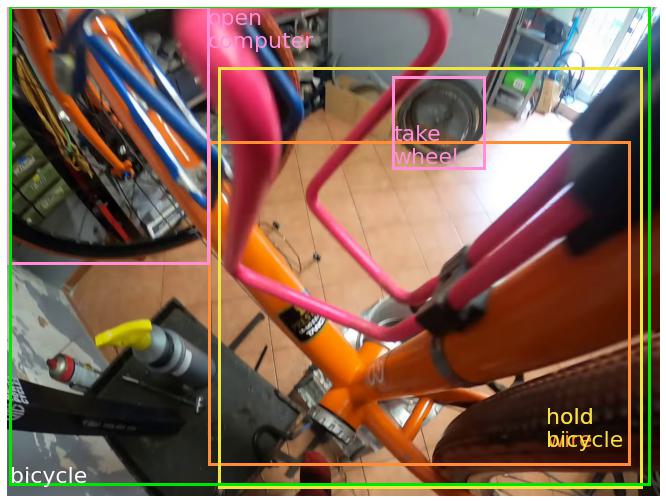}
      &
    \includegraphics[width=.3\textwidth]{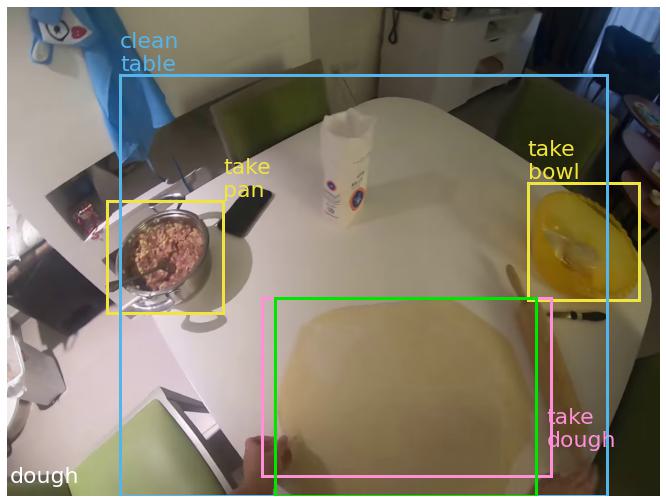} \\
    
    \includegraphics[width=.3\textwidth]{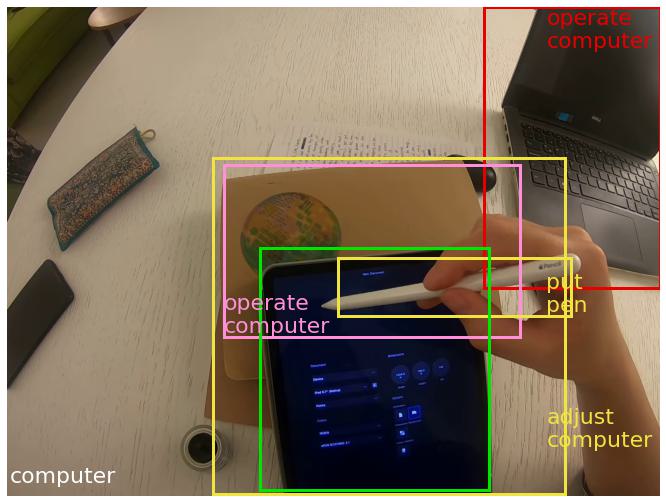}
     &
      \includegraphics[width=.3\textwidth]{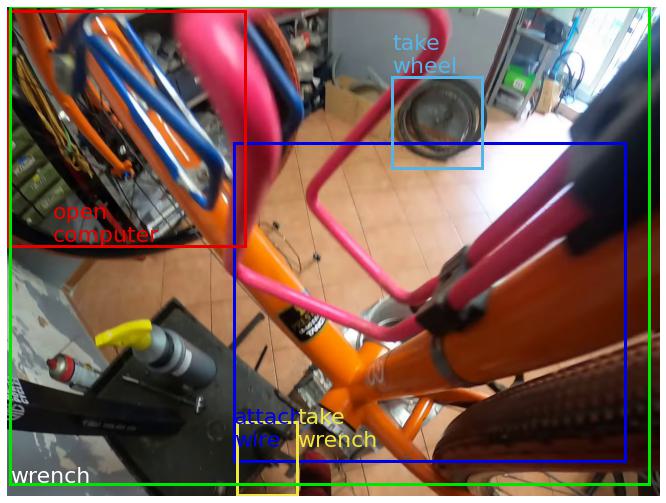}
      &
    \includegraphics[width=.3\textwidth]{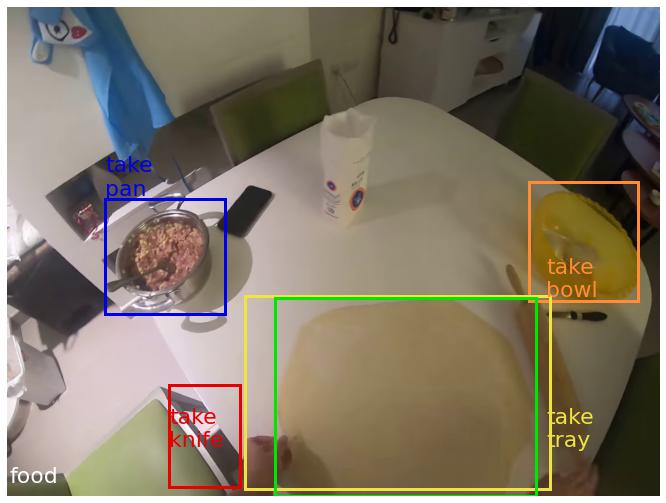} \\
        \end{tabular}
        \caption{\centering{\textbf{Additional qualitative examples of predictions when changing the language input}. Our model modifies the predicted labels and locations \textit{dynamically} based on the input language context descriptions. We show the input language context in white, in the bottom left corner of each image.}
        \label{fig:language_change_extra}}
\end{figure*}

\begin{figure*}[!ht]
    \centering
    \begin{tabular}{@{\hspace{0.75\tabcolsep}} c @{\hspace{0.75\tabcolsep}} c @{\hspace{0.75\tabcolsep}} c}
      GT (red box), $\mathcal{N}^F_h$ obj. (blue box) & Hand interaction detector output & Object detector output \vspace*{0.2cm} \\ 
      
     \includegraphics[width=.31\linewidth]{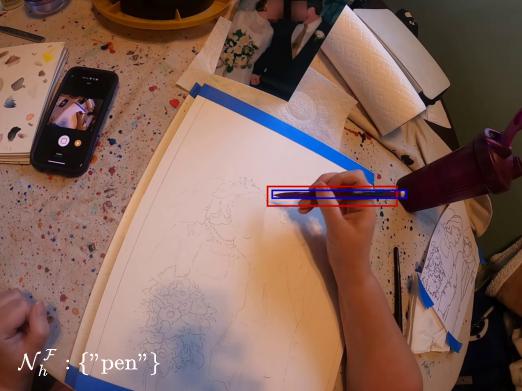} 
     & \includegraphics[width=.31\linewidth]{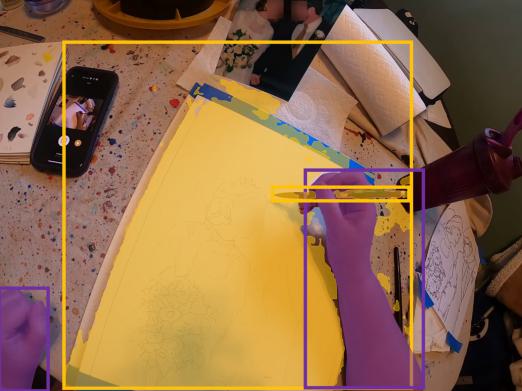} 
     & \includegraphics[width=.31\linewidth]{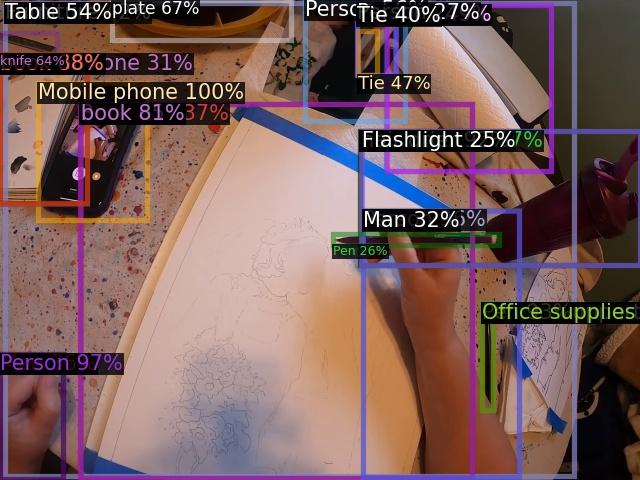}
     \vspace*{0.1cm}
     \\

    \includegraphics[width=.31\linewidth]{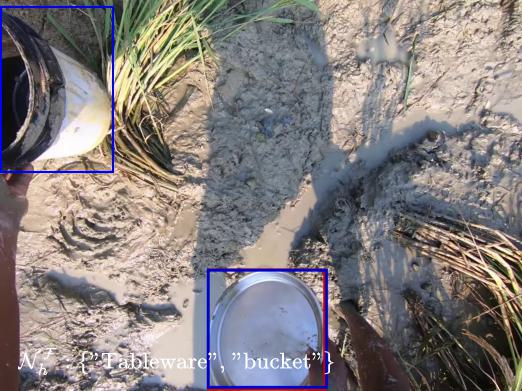} 
    & \includegraphics[width=.31\linewidth]{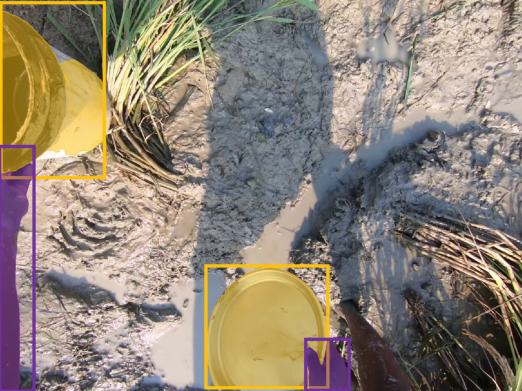}
    & \includegraphics[width=.31\linewidth]{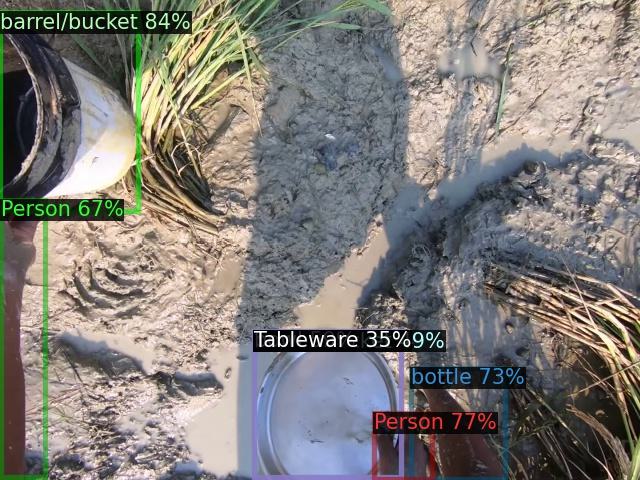} 
    \vspace*{0.1cm}\\      
    
      \includegraphics[width=.31\linewidth]{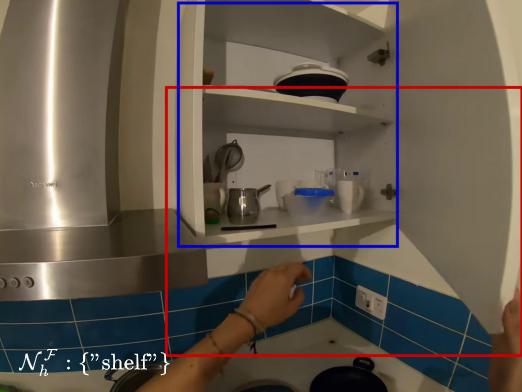}
     & \includegraphics[width=.31\linewidth]{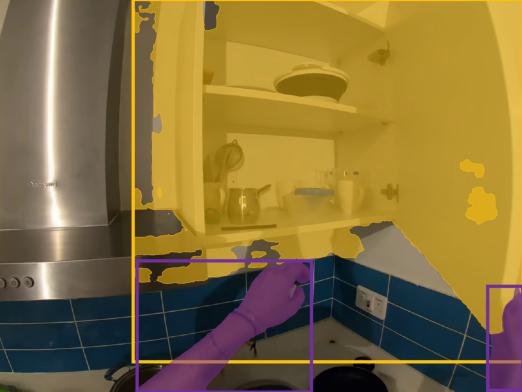} 
     & \includegraphics[width=.31\linewidth]{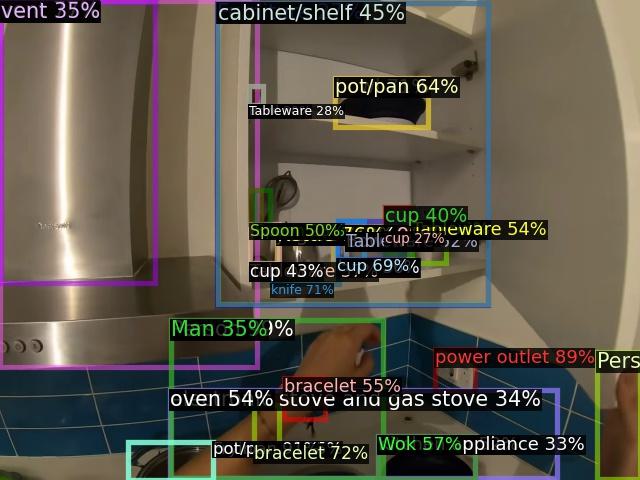}
     \vspace*{0.1cm} \\
            
     \includegraphics[width=.31\linewidth]{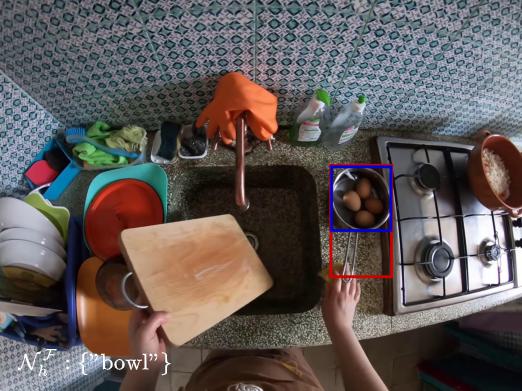}
    & \includegraphics[width=.31\linewidth]{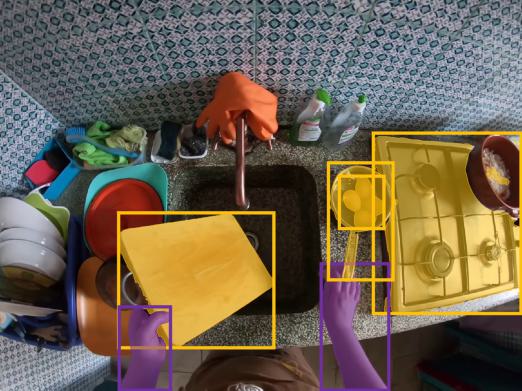} & \includegraphics[width=.31\linewidth]{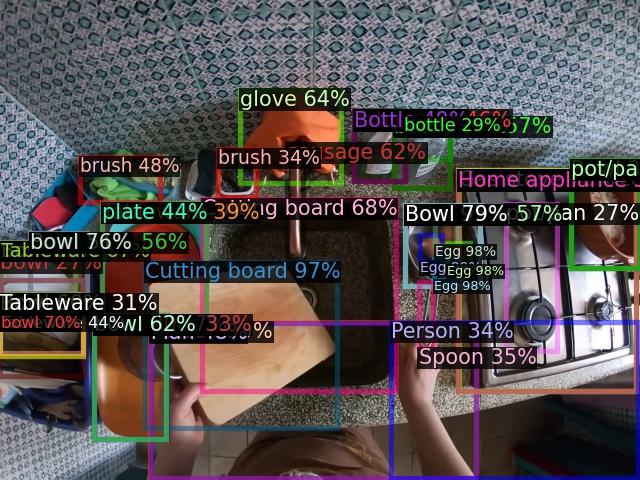} 
    \\%
    \end{tabular}

    \caption{
    \centering{\textbf{Successful detections of held objects for $\mathcal{N}_h$ construction.} We show the ground-truth next-active object in red, hand bounding boxes in purple, active object bounding boxes in yellow, the bounding boxes of the inferred held objects, selected from those detected by the object detector and having a sufficiently high IoU with an active object bounding box from the hand interaction model, in blue, and the extracted held object context $\mathcal{N}^F_h$ in the bottom left corner of each image.}}
   \label{sup_fig:n_h_good}
\end{figure*}

\begin{figure*}[!ht]
    \centering
    \begin{tabular}{@{\hspace{0.75\tabcolsep}} c @{\hspace{0.75\tabcolsep}} c @{\hspace{0.75\tabcolsep}} c}
      GT (red box) & Hand interaction detector output & Object detector output \vspace*{0.2cm} \\
            
     \includegraphics[width=.31\linewidth]{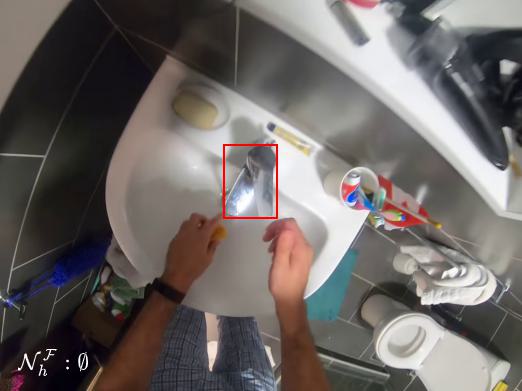} 
     & \includegraphics[width=.31\linewidth]{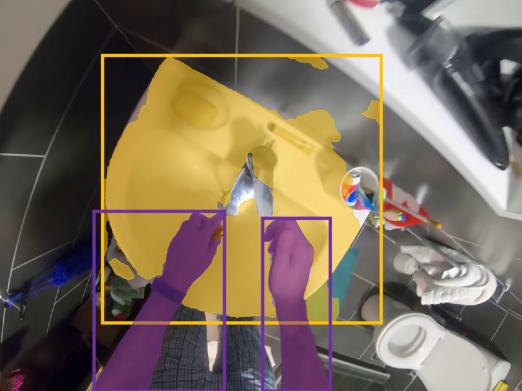} 
     & \includegraphics[width=.31\linewidth]{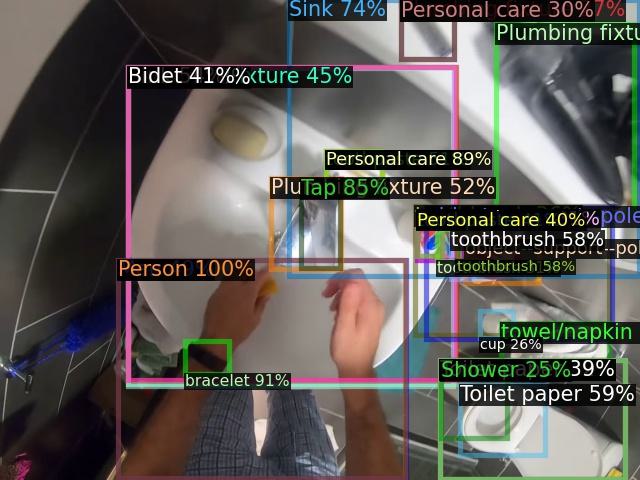}
     \vspace*{0.1cm}
     \\

    \includegraphics[width=.31\linewidth]{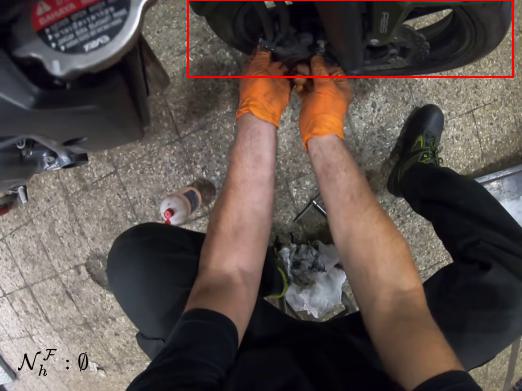} 
    & \includegraphics[width=.31\linewidth]{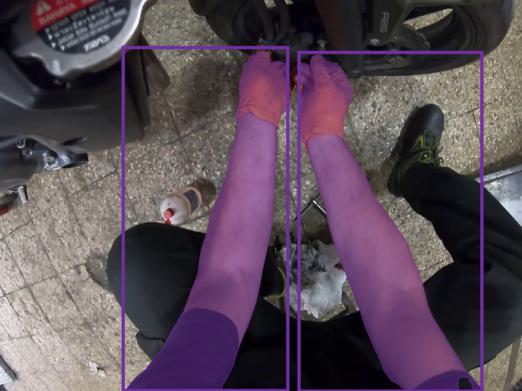}
    & \includegraphics[width=.31\linewidth]{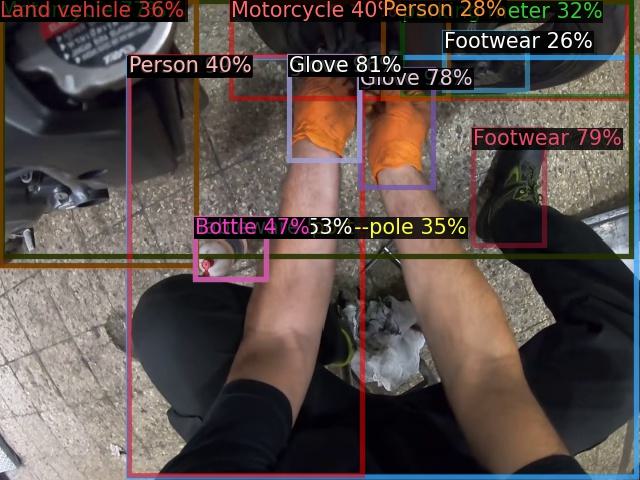} \\

    \end{tabular}

    \caption{
    \centering{\textbf{Failures of hand-object interaction detector during $\mathcal{N}_h$ construction}. The hand-object interaction detector produces an oversegmentation (top) or does not detect the object (bottom). See \autoref{sup_fig:n_h_good} for details.}}
   \label{sup_fig:n_h_visor_fail}
\end{figure*}

\begin{figure*}[!ht]
    \centering
    \begin{tabular}{@{\hspace{0.75\tabcolsep}} c @{\hspace{0.75\tabcolsep}} c @{\hspace{0.75\tabcolsep}} c}
      GT (red box), $\mathcal{N}^F_h$ obj. (blue box) & Hand interaction detector output & Object detector output \vspace*{0.2cm} \\
           
    \includegraphics[width=.31\linewidth]{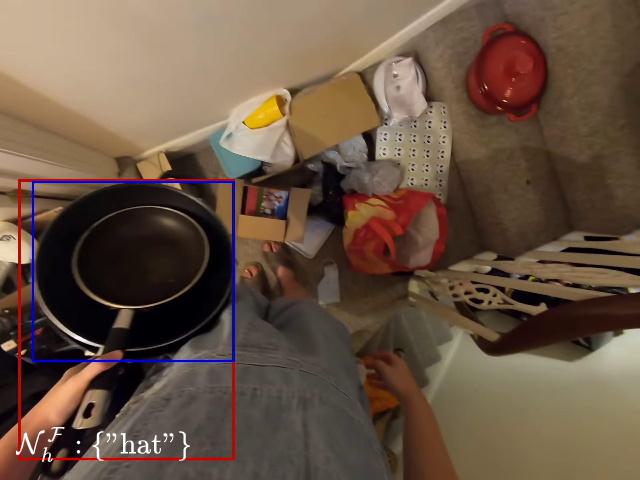} 
    & \includegraphics[width=.31\linewidth]{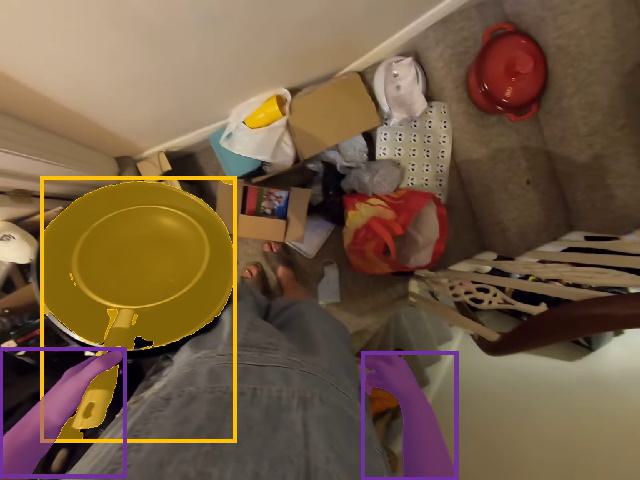}
    & \includegraphics[width=.31\linewidth]{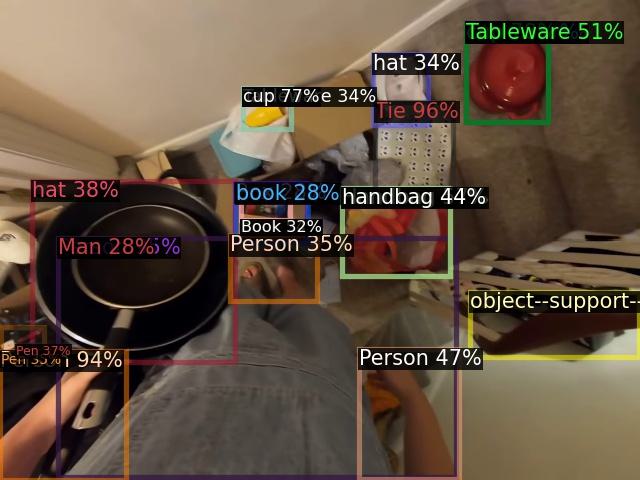} 
    \vspace*{0.1cm}\\
          
     \includegraphics[width=.31\linewidth]{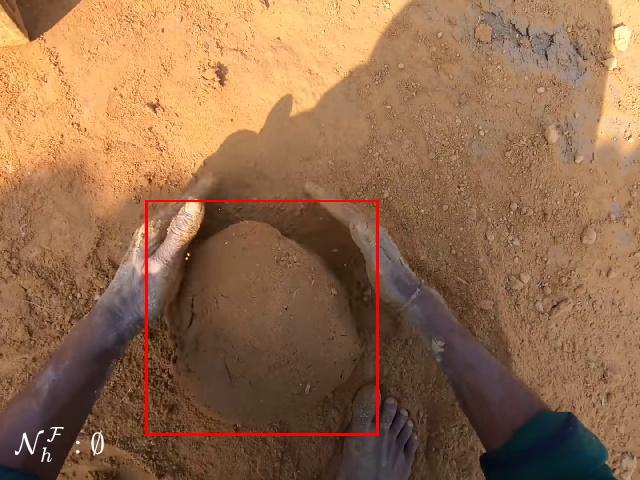} 
     & \includegraphics[width=.31\linewidth]{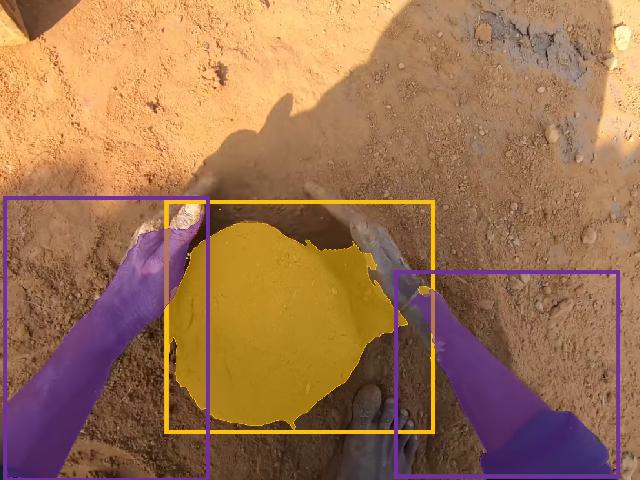} 
     & \includegraphics[width=.31\linewidth]{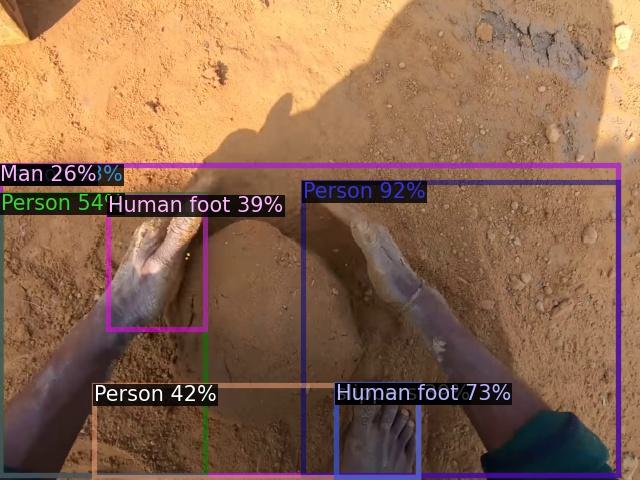} \\
    \\

    \end{tabular}

    \caption{
    \centering{\textbf{Failures of object detector during $\mathcal{N}_h$ construction}. The object is misclassified (top) or not detected (bottom). See \autoref{sup_fig:n_h_good} for details.}}
   \label{sup_fig:n_h_unidet_fail}
\end{figure*}

We present in Figure \ref{sup_fig:qual_1} and Figures \ref{sup_fig:qual_2} \& \ref{sup_fig:qual_3} a qualitative comparison between our method and that of Ego4D, using a context length of 3 for action verb-noun pairs $\mathcal{A}$ and 3 salient objects $\mathcal{N}_s$ for our language-aided model. The green bounding box represents the ground truth location of the next object interaction. The white text represents the input action summary context description: the first row represents the salient objects and the second the captioned action descriptions. We show the top-4 most confident bounding box predictions and their associated noun, verb, and TTC estimates.

\end{document}